\UseRawInputEncoding
\documentclass[11pt]{article}
\usepackage{amsfonts}
\usepackage{}
\usepackage{mathrsfs}
\usepackage{amssymb}
\usepackage{amsmath}
\usepackage{cite}
\usepackage{color}
\usepackage{extarrows}
\usepackage{graphicx}
\usepackage{subfigure}
\usepackage{float}
\usepackage{diagbox}
\usepackage{booktabs}
\usepackage{algorithm}
\usepackage{algpseudocode}
\usepackage{caption}
\allowdisplaybreaks[4]

\def\no{\nonumber}

\newcommand\btd{\raise 2pt \hbox{$\hat\bigtriangledown$}\hskip 1.5pt}
\newcommand\bt{\raise 2pt \hbox{$\bigtriangledown$}\hskip 1.5pt}

\def\no{\nonumber}
\def\x{\textbf{x}}

\def\b{\textbf{b}}
\def\w{\textbf{w}}

\begin{document}
\title{{Enforcing continuous symmetries in physics-informed  neural network for solving forward and inverse problems of partial differential equations }}
\author{Zhi-Yong Zhang $^1$ \footnote{E-mail: zzy@muc.edu.cn (Corresponding author)}\ \ \ Hui Zhang $^1$ \footnote{E-mail: zh13276358263@163.com}\ \ \ Li-Sheng Zhang $^2$ \footnote{E-mail: zls@ncut.edu.cn}\ \ \  Lei-Lei  Guo $^2$ \footnote{E-mail: leiguo@mmrc.iss.ac.cn}\ \ \ \ \ \
 \\
\small $^1$ College of Science, Minzu University of China, Beijing 100081, P.R. China\\
\small$^2$ College of Science, North China University of Technology, Beijing 100144, P.R. China}
\date{}
\maketitle

\noindent{\bf Abstract:} As a typical application of deep learning, physics-informed neural network (PINN) {has been} successfully used to find numerical solutions of partial differential equations (PDEs), but how to improve the limited accuracy is still a great challenge for PINN. In this work, we introduce a new method, symmetry-enhanced physics informed neural network (SPINN) where the invariant surface conditions induced by the Lie symmetries or non-classical symmetries of PDEs are embedded into the loss function in PINN, to improve the accuracy of PINN for solving the forward and inverse problems of PDEs. We test the effectiveness of SPINN for the forward problem via two groups of ten independent numerical experiments using different numbers of collocation points and neurons for the heat equation, Korteweg-de Vries (KdV) equation and potential Burgers {equations} respectively, and for the inverse problem by considering different layers and neurons as well as different training points for the Burgers equation in potential form. The numerical results show that SPINN performs better than PINN with fewer training points and simpler architecture of neural network. Furthermore, we discuss the computational overhead of SPINN in terms of the relative computational cost to PINN and show that the training time of SPINN has no obvious increases, even less than PINN for certain cases.


\section{Introduction}
Partial differential equations (PDEs) describe the complex phenomenon in various fields such as physics, chemistry and biology, thus finding solutions of PDEs is of great interest and a direct and effective way to study the dynamical behaviors of PDEs. Among the existing methods, the first choice is to use the approaches related with intrinsic properties of PDEs to construct exact solutions. The classical Lie symmetry theory of PDEs is one of the main inspiring sources for various new methods to obtain exact solutions for PDEs and also very useful to detect the integrability, translational or rotational invariance, and to construct conservation laws of PDEs \cite{olv,blu}. Specifically, a continuous Lie symmetry of PDEs is a continuous transformation which maps one solution to another solution of the same PDEs. Furthermore, one can construct invariant solutions of PDEs via the invariant surface conditions generated by the Lie symmetries and also perform symmetry reduction technique to reduce the PDEs into the ones with fewer independent variables. However, for some symmetry reduced PDEs, it is still difficult to find explicit exact solutions with the known methods, thus one resorts to numerical methods to search for numerical solutions of PDEs.

In addition to the traditional sophisticated numerical methods, such as finite element, finite difference and finite volume, the physics-informed neural networks (PINN) in deep learning field attracted more attentions in the data-driven discovery of solutions of PDEs in recent years \cite{2018a}. The core idea of PINN is to represent the solutions of PDEs by a neural network where the parameters are trained via gradient descent of the loss function {related with the PDEs as well as the initial and boundary conditions. In particular, the technique of automatic differentiation is employed by the deep learning community to deal with the derivatives \cite{auta}. }Up to now, PINN has been widely applied in the {field }of scientific computing, especially in solving forward and inverse problems of nonlinear PDEs due to the merits of flexibility and gridless nature \cite{yan-2021,chen-2021,lmz-2021}.
The physics in PINN is described by the considered PDEs. However, the solutions may fail the physical properties of the equations such as the Lie symmetries and conservation laws. Thus in order to address this gap, the improved versions of PINN spring up. For example, a gradient-enhanced PINN was proposed in solving both forward and inverse problems of PDEs where the gradient information of the PDEs residual is embed into the loss function \cite{J}. Lin and Chen introduced a two-stage PINN where the first stage is the PINN and the second stage is to incorporate the conserved quantities into mean squared error loss to train neural networks \cite{jan}. In \cite{tb-2022}, the authors enforced nonlinear analytic constraints in the architecture of neural network {into} the loss function to produce the results consistent with the constraints. Zhu et.al constructed the group-equivariant neural networks which respect the spatio-temporal parity symmetries and successfully emulated different types of periodic solutions of nonlinear dynamical lattices \cite{zhu-2022}. Jagtag et al. proposed local adaptive activation functions by introducing a scalable parameter in each layer and for every neuron  separately to accelerate the training speed and convergence \cite{ja-2020}. In particular, some clever techniques such as the properly-designed non-uniform training point weighting \cite{sy-2020}, Domain decomposition \cite{vb-2020} and digging a-priori information of solutions of PDEs \cite{wls-2020} are very effective for improving the accuracy and efficiency of PINN.
More targeted neural networks such as the Bayesian PINN \cite{yang-2021}, discrete PINN framework based on graph convolutional network and variational structure of PDE \cite{gao-2022}, extended physics-informed neural networks (XPINNs) \cite{jag-2020}, parareal physics-informed neural network (PPINN) \cite{meng-2020}, auxiliary physics informed neural networks (A-PINN) \cite{ly-2022}, fractional physics-informed neural networks (fPINNs) \cite{fpinn-2019} and variational PINN \cite{ekh-2021} were devised to eliminate roadblocks in more complex and realistic applications. Essentially, the deep neural network for solving PDEs is an optimal fitting operation on a limited data set, thus the inherent propertied of PDEs such as the Lie symmetry must be benefit for the procedure.

In this paper, we propose a new method, PINN enforced by the continuous Lie symmetry information of PDEs (SPINN), where the invariant surface conditions (ISC) induced by the Lie symmetries are incorporated into the loss function in PINN. If the PDEs together with the initial and boundary conditions admit a Lie symmetry, then the solutions are invariant under the Lie symmetry and also satisfy the ISC, i.e. the ISC are the inherent properties of the PDEs and place new essential constraints on the solutions.
Since the standard loss function in PINN is the mean square error of PDE residual, thus adding the ISC to the loss function will
{definitely increase new constraints for the objective function during the optimization process and improve the accuracy of the solutions}. Our method is fundamentally different from those prior approaches which use parity-symmetry in the framework of neural networks \cite{zhu-2022} or the information of conservation law in the loss function \cite{jan}, and thus the first time to embed the continuous symmetries into the loss functions of PINN. Furthermore, we perform two groups of ten independent experiments for the three PDEs respectively and show that SPINN largely outperforms than PINN, even in the cases where SPINN performs poorly the experiment result is still better than the corresponding one of PINN under the same training data set and initializations. In addition, the symmetries admitted by the PDEs together with the initial and boundary conditions are further studied in \cite{blu,zhang-2010,goard-2008} and thus the proposed SPINN has wide application prospects. It is worthy of saying that the non-classical symmetry, which was first introduced by Bluman and Cole to obtain new exact solutions of the linear heat equation,  is also effective for generating ISC \cite{blu-1969}.

The remainder of the paper is arranged as follows. In the following section, we first briefly review the main idea of PINN and the framework of Lie symmetry of PDEs, and introduce SPINN in detail. In Section 3, two groups of ten independent experiments are performed for the heat equation, KdV equation and the potential Burgers equation to illustrate the effectiveness of SPINN. The computational cost of SPINN is also discussed and demonstrates that the error accuracy of the numerical solutions by SPINN is greatly improved without obvious increase of training time. In Section 4, we use SPINN to study the inverse problem of PDEs which is exemplified by the Burgers equation in potential form. A systematic investigation of SPINN with different network structures and different training points with different noise levels is considered and the training time of SPINN is also analysed via the relative computational cost to PINN.  We conclude the results in the last section.
\section{Main ideas of the methods}

We take the following system of two $r$-th order PDEs
\begin{eqnarray} \label{eqn1}
&&\no f: =u_t+\mathcal {N}_u[u,v]=0,\\
&& g: =v_t+\mathcal {N}_v[u,v]=0,~~~~~t\in [0, T],~~x\in\Omega,
\end{eqnarray}
together with the initial and boundary conditions
\begin{eqnarray}\label{ib}
&&\no I(x,u,v)=0;\\
&& \mathcal {B}(t,u,v)=0,~~~~ \mbox{on} ~~~\partial \Omega,
\end{eqnarray}
as an example to introduce the main ideas of the methods, where $u=u(t,x)$ and $v=v(t,x)$ are the solutions to be determined, $\Omega$ denotes a finite interval, $\mathcal {N}_u[u,v]$ and $\mathcal {N}_v[u,v]$ denote the smooth functions of $u,v$ and their $x$-derivatives up to $r$-th order. 

\subsection{PINN for solving PDEs}
The main idea of PINN for solving PDEs is to construct a neural network $(\widetilde{u}(t,x,\Theta),\widetilde{v}(t,x,\Theta))$ to approximate the exact solutions $(u(t,x),v(t,x))$ via the trainable parameters $\Theta$. Specifically, consider a $K$ layers neural network
consisting of one input layer, $K-1$ hidden layers and one output layer. The $k$th $(k=1,2,\dots,K)$ layer has $N_k$
neurons, which means that it transmits $N_k$-dimensional output vector $\x_l$ to the $(k + 1)$th layer as the input data. The
connection between two layers is built by the linear transformation $\mathcal {F}_k$ and the nonlinear activation function $\sigma(\cdot)$:
\begin{eqnarray}
&&\no \x_k = \sigma(\mathcal {F}_k(\x_{k-1})) = \sigma(\w_k\x_{k-1} + \b_k),
\end{eqnarray}
where $\w_k\in \mathbb{R}^{N_k\times N_{k-1}}$ and $\b_k\in \mathbb{R}^{N_k}$ denote the weight matrix and bias vector of the $k$th layer respectively. Thus, the connection
between the input $\x_0$ and the output $u(\x_0, \Theta)$ is given by
\begin{eqnarray}
&&\no u(\x_0, \Theta) = (\mathcal {F}_k\circ \sigma \circ \mathcal {F}_{k-1} \circ \cdots\circ \sigma \circ \mathcal {F}_1)(\x_0),
\end{eqnarray}
and $\Theta=\{\w_k,\b_k\}_{k=1}^K$ is the trainable parameters of PINN. The most frequently used activation function in the neural network for solving PDEs is the hyperbolic tangent (tanh) function while the weights and bias are initialized by the Xavier initialization and the derivatives of $(u(t,x),v(t,x))$ with respect to time $t$ and space $x$ are derived by automatic differentiation \cite{2018a}.  Meanwhile, PINN method usually utilizes the Adam \cite{km-2015} or the L-BFGS \cite{ln-1989} algorithms to update the parameters $\Theta$ and thus to minimize the loss function of mean square error (MSE)
\begin{eqnarray} \label{invaloss}
&&MSE=w_bMSE_b+w_iMSE_i+w_fMSE_f+w_gMSE_g,
\end{eqnarray}
where both $MSE_i$ and $MSE_b$ work on the initial and boundary data while $MSE_f$ and $MSE_g$ enforce the structure imposed by system (\ref{eqn1}) at a finite set of collocation points,
\begin{eqnarray}\label{residual}
&&\no MSE_b=\frac{1}{N}\sum_{i=1}^{N}{\mid\mathcal {B}(t^i,u^{i}, v^{i})\mid}^{2},\\
&&\no MSE_i=\frac{1}{N}\sum_{i=1}^{N}{\mid I(x^i,u^{i}, v^{i})\mid}^{2},\\
&& MSE_f=\frac{1}{\widetilde{N}}\sum_{j=1}^{\widetilde{N}}{\mid f(\widetilde{t}_{j},\widetilde{x}_{j})\mid}^{2},~~~MSE_g=\frac{1}{\widetilde{N}}\sum_{j=1}^{\widetilde{N}}{\mid g(\widetilde{t}_{j},\widetilde{x}_{j})\mid}^{2},
\end{eqnarray}
where $w_b,w_i, w_f$ and $w_g$ are the weights, $\left\{t^{i}, x^{i}, u^{i}, v^{i}\right\}_{i=1}^N$ are the initial and boundary training data and $\left\{\widetilde{t}_{j}, \widetilde{x}_{j}\right\}_{j=1}^{\widetilde{N}}$ denote the collocation points for $f$ and $g$.

PINN is also effective for the inverse problem of PDEs where the undetermined parameters as well as the numerical solutions are learned together. If there exist some unknown parameter $\lambda$ in Eq.(\ref{eqn1}), then one can learn them by adding the extra
measurements of $(u(t,x),v(t,x))$ on the set of training points $T_i$, i.e. an additional data loss [3,4] as
\begin{eqnarray}
&&\no MSE_p=\frac{1}{\widetilde{N}}\sum_{j=1}^{\widetilde{N}}\left[\,|\widetilde{u}(\widetilde{t}_{j},\widetilde{x}_{j})-u^i|^2
+|\widetilde{v}(\widetilde{t}_{j},\widetilde{x}_{j})-v^i|^2\,\right],
\end{eqnarray}
thus the new loss function for  the inverse problem of PDEs is given by
\begin{eqnarray} \label{invaloss-inv}
&&MSE=w_pMSE_p+w_bMSE_b+w_iMSE_i+w_fMSE_f+w_gMSE_g.
\end{eqnarray}
Note that in this study we choose the weights $w_p=w_b=w_i=w_f=w_g=1$ in both PINN and the SPINN below, thus in SPINN we do not state them again.

Generally speaking, the total number of training {datas} $N$ is relatively small (a few hundred up to a few thousand points) while the number of collocations points are large.  Alternatively, one can construct a new network by enforcing the boundary or initial conditions satisfied exactly and automatically and thus eliminate the loss term of boundary conditions \cite{pll-2020,lu-2021}. Such a method generally has a more higher accuracy than the network whose loss function contains the residuals of initial and boundary conditions. Here we still adopt the loss function in (\ref{invaloss}) and thus the proposed SPINN below may have more higher accuracy by the alternative method.
\subsection{Symmetry of PDEs}
Symmetry is an inherent but not exposed property of PDEs. It provides widely applicable approach to find closed form solutions of PDEs where most of the useful systematic methods for solving PDEs such as separation of variables involve direct uses of symmetry method.
The classical method for obtaining Lie symmetry admitted by system (\ref{eqn1}) is to find a local one-parameter Lie group of infinitesimal transformation
\begin{eqnarray}\label{group}
&&\no x^*= x+\varepsilon\,\xi (x,t,u,v)+O(\varepsilon^2),\\
&&\no t^*=t+\varepsilon\,\tau (x,t,u,v)+O(\varepsilon^2),\\
&&\no u^*=u+\varepsilon\,\eta (x,t,u,v)+O(\varepsilon^2),\\
&& v^*=v+\varepsilon\,\phi (x,t,u,v)+O(\varepsilon^2),
\end{eqnarray}
which leaves system (\ref{eqn1}) invariant. Lie's fundamental theorem shows that such a group is completely characterized by the
infinitesimal operator
$ \mathcal {X}=\xi\partial_x+\tau\partial_t+\eta\partial_u+\phi\partial_v$, where we briefly denote $\xi=\xi(x,t,u,v),\tau=\tau(x,t,u,v),\eta=\eta(x,t,u,v)$ and $\phi=\phi (x,t,u,v)$, thus we will not differentiate the Lie group (\ref{group}) and the corresponding operator $\mathcal {X}$ and call them as Lie symmetry.
Then Lie's infinitesimal criterion requires $\mathcal {X}$ satisfying
\begin{equation}\label{deter-1}
\no  \text{pr}^{(2)}\mathcal {X}(f)_{|\{(\ref{eqn1})\}}=\text{pr}^{(2)}\mathcal {X}(g)_{|\{(\ref{eqn1})\}}=0,
\end{equation}
where, here and below, the symbol $_{|\{\Delta\}}$ means that the computations work on the solution space of $\Delta=0$, and
 \begin{eqnarray}
&&\no \text{pr}^{(2)}\mathcal {X}=\mathcal {X}+\eta_t^{(1)}\partial_{u_t}+\eta_x^{(1)}\partial_{u_1}+\eta_x^{(2)}\partial_{u_2}+\dots+\eta_x^{(r)}\partial_{u_r}\\
&&\hspace{2cm} +\phi_t^{(1)}\partial_{v_t}+\phi_x^{(1)}\partial_{v_1}+\phi_x^{(2)}\partial_{v_2}+\dots+\phi_x^{(r)}\partial_{v_r}
\end{eqnarray}
stands for $r$-th order prolongation of $\mathcal {X}$, $u_i=\partial^i u/\partial x^i$ with $i=1,2,\dots,r$, the coefficients $\eta_t^{(1)},\eta_x^{(i)}$ and $\phi_t^{(1)},\phi_x^{(i)}$ can be calculated by the well-known prolongation formulae \cite{blu,olv},
 \begin{eqnarray}\label{formula}
&&\no \eta_t^{(1)}=D_t\left(\eta-\tau u_t-\xi\,u_x\right)+\tau D_t^2u+\xi\,D_xD_tu,\\
&&\no \eta_x^{(i)}=D_x^i\left(\eta-\tau u_t-\xi\,u_x\right)+\tau D_tD_x^iu+\xi\,D_x^{i+1}u,\\
&&\no \phi_t^{(1)}=D_t\left(\phi-\tau v_t-\xi\,v_x\right)+\tau D_t^2v+\xi\,D_xD_tv,\\
&& \phi_x^{(i)}=D_x^i\left(\phi-\tau v_t-\xi\,v_x\right)+\tau D_tD_x^iv+\xi\,D_x^{i+1}v.
\end{eqnarray}
The symbols $D_t$ and $D_x$ in (\ref{formula}) indicate the total derivatives with respect to $t$ and $x$ respectively,
 \begin{eqnarray}
&&\no D_t=\partial_t+u_t\partial_u+u_{xt}\partial_{u_{x}}+u_{tt}\partial_{u_t}+v_t\partial_v+v_{xt}\partial_{v_{x}}+v_{tt}\partial_{v_t}+\dots,\\
&&\no D_x=\partial_{x}+u_1\partial_u+u_2\partial_{u_1}+u_{xt}\partial_{u_t}+v_1\partial_v+v_2\partial_{v_1}+v_{xt}\partial_{v_t}+\dots,
\end{eqnarray}
and $D_t^0(u)=u$, {$D_t^{r}=D_t\left(D_t^{r-1}\right)$ }and similar for $D_x$.

The Lie symmetry admitted by (\ref{eqn1}) can be used to construct {similarity} solutions which also satisfy the ISC
 \begin{eqnarray}\label{isc}
&&\no ISC_1:=\eta-\tau u_t-\xi\,u_x=0,\\
&& ISC_2:=\phi-\tau v_t-\xi\,v_x=0.
\end{eqnarray}

Hence, the ISC (\ref{isc}) add new constraints for finding solutions of system (\ref{eqn1}) if the system admits the Lie symmetry (\ref{group}). Moreover, the methods for finding symmetries of system (\ref{eqn1}) with initial-boundary value conditions (\ref{ib}) are well-studied \cite{blu,zhang-2010,goard-2008}, thus it is feasible to use the ISC to improve the accuracy of learned solutions by deep neural network.

If the system (\ref{eqn1}) is compatible with the ISC (\ref{isc}), then on the common solution manifold of the two systems, the conditions
\begin{equation}\label{deter-non}
\no \text{pr}^{(2)}\mathcal {X}(f)_{|\{(\ref{eqn1}),(\ref{isc})\}}=\text{pr}^{(2)}\mathcal {X}(g)_{|\{(\ref{eqn1}),(\ref{isc})\}}=0,
\end{equation}
generate the non-classical symmetry which obviously contains Lie symmetry as a particular case \cite{blu-1969}. The non-classical symmetry also keeps the graph of common solutions of systems (\ref{eqn1}) and (\ref{isc}) invariant and thus exerts the same role of Lie symmetry to learn high accuracy solutions by deep neural network.
\subsection{SPINN for solving PDEs}
In addition to considering PDEs, there exist no `physical' elements expressed in PINN \cite{2018a}. However, the  predicted solutions of PDEs via the PINN contain the physical properties, such as the Lie symmetry or non-classical in subsection 2.2, which are not reflected in the training procedure of neural networks.
Observe that the solutions of the initial boundary problem associated with system (\ref{eqn1}) are also restricted by the ISC, thus it is reasonable to embed the ISC into the loss functions of PINN to further improve the efficiency of neural network. Meanwhile, the considered symmetries, Lie symmetry or non-classical symmetry, characterize the inherent properties of PDEs and thus their induced ISC can accelerate the learning efficiency.

Specifically, the loss function in SPINN involves the constraints of the ISC in addition to the one in PINN and takes the form
\begin{eqnarray} \label{inva-sym}
&&MSE=MSE_i+MSE_b+MSE_f+MSE_g+MSE_{ISC},
\end{eqnarray}
where $MSE_i, MSE_b$ and $MSE_f, MSE_g$ are defined in (\ref{residual}) and
\begin{eqnarray}
&&\no MSE_{ISC}=\frac{1}{\widetilde{N}}\sum_{i=1}^{\widetilde{N}}\big(\mid ISC_1(\widetilde{t}_{j}, \widetilde{x}_{j})\mid^{2}+ \mid ISC_2(\widetilde{t}_{j}, \widetilde{x}_{j})\mid^{2}\big).
\end{eqnarray}

\begin{figure}[htp]
\centering
	\begin{minipage}{0.88\linewidth}
		\vspace{3pt}
		\centerline{\includegraphics[width=\textwidth]{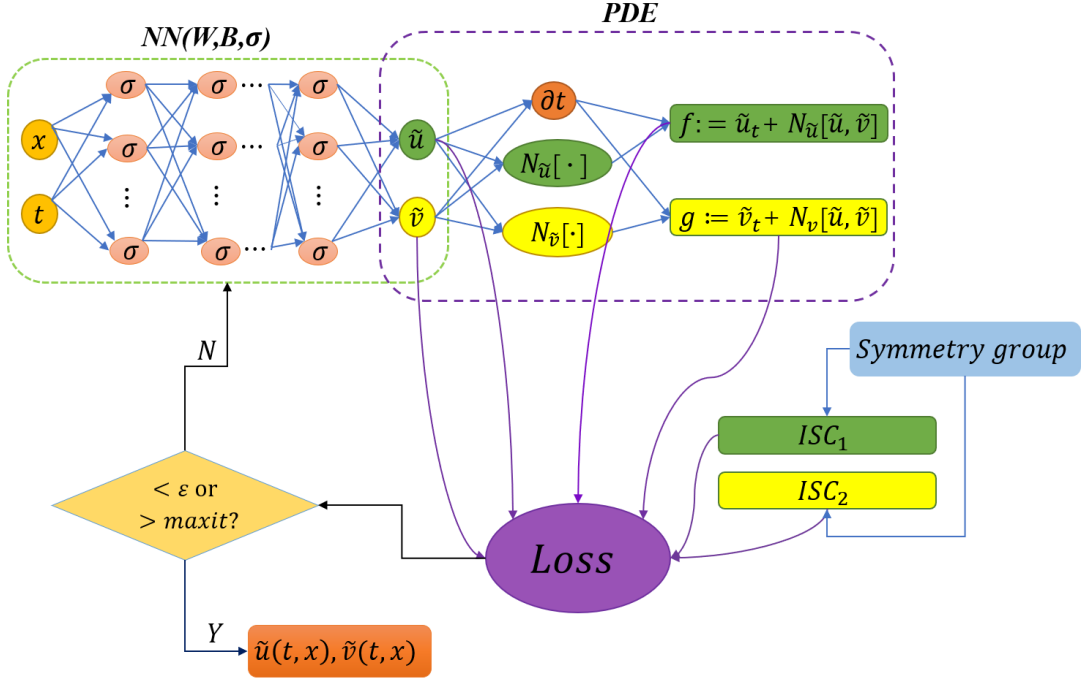}}
	\end{minipage}
	\caption{(Color online) Schematic diagram of SPINN, where maxit denotes the maximum number of iterations.}
\label{fig1-flow}
\end{figure}

We state the whole procedure of SPINN in Algorithm 1 and depict the schematic diagram in Figure \ref{fig1-flow} to clarify the whole procedure of SPINN.
\begin{algorithm}[htbp]
\caption{: Steps for SPINN}
  \label{alg:Framwork}
  \begin{algorithmic}[1]
    \Require
      Initial and boundary data set: $\left\{t^{i}, x^{i}, u^{i}, v^{i}\right\}_{i=1}^N$;
      Initial value of weights and bias: Xavier initialization;
      Loss functions: MSE in (\ref{inva-sym});
    \Ensure
      Learned solutions: $\big(\widetilde{u},\widetilde{v}\big)$ and $MSE$;
    \State Find the Lie symmetry or non-classical symmetry of PDEs:  $\mathcal {X}$;
    \label{code:fram:extract}
    \State Get the ISC according the obtained the Lie symmetry or non-classical symmetry of PDEs:  $ISC_1$ and $ISC_2$;
    \State Train the neural network with the loss function given in (\ref{inva-sym});
    \label{code:fram:trainbase} \\
    \Return the error and predicted solution $\big(\widetilde{u},\widetilde{v}\big)$ satisfying the given error bound;
  \end{algorithmic}
\end{algorithm}

Compared with PINN, the optimization problem in SPINN is to find the minimum value of the loss function, composed of PDEs residual and the ISC residual, by optimizing the weights and biases. As we will show in the numerical experiments, SPINN improves the accuracy of predicted solutions, and requires less training points and  simpler networks to achieve the accuracy of PINN with more training points. 

\section{Numerical experiments}
In this section, we apply the proposed SPINN to study three PDEs: KdV equation with the time and space translation symmetry, Heat equation with Lie symmetry and the potential Burgers equation with non-classical symmetry. In the three examples, we choose the tanh as the activation function and use the L-BFGS algorithm, a full batch gradient descent optimization algorithm based on the quasi-Newton method \cite{ln-1989}, to optimize the loss function to gather {the $L_2$ norm error}.

For each example, keep all other elements of neural network and data unchanged, we perform ten independent experiments for two cases respectively: the changes of collocation points and the changes of numbers of neurons. The experiment results show that SPINN outperforms than PINN with the same architecture and parameters.

\subsection{KdV equation}
The nonlinear KdV equation reads as
\begin{eqnarray} \label{kdv}
&& u_t+uu_x+u_{xxx}=0,~~~~x\in[0,1],~~t\in [0,1],
\end{eqnarray}
which arises in the theory of long waves in shallow water and the physical systems in which both nonlinear and dispersive effects are relevant \cite{kdv-1894}, and the initial boundary conditions are
\begin{eqnarray}
&&\no u(0,x)=12\mathrm{sech}^{2} x,\\
&&\no u(t,0)=12 \mathrm{sech}^{2} (-4t),~~~u(t,1)=12 \mathrm{sech}^{2} (1-4t).
\end{eqnarray}

Eq.\eqref{kdv} is admitted by the combination of time and space translation symmetries $\mathcal {X}_{kdv}=\partial_t+c\,\partial_x$, where $c=4$ is called the speed of travelling wave. By means of the symmetry $\mathcal {X}_{kdv}$, we get the celebrated soliton solution $u=12 \mathrm{sech}^2(x-4t)$ and the ISC $u_t+4u_x=0$.
 Let
\begin{eqnarray} \label{lieback}
&&\no f:=u_t+uu_x+u_{xxx},\\
&&\no g:=u_t+4u_x.
\end{eqnarray}
Then the shared parameters of the neural networks $(u,f,g)$ with SPINN can be learned by minimizing the mean squared error
\begin{eqnarray} \label{infini2}
&& MSE=MSE_u+MSE_f+MSE_g,
\end{eqnarray}
where
\begin{eqnarray}
&&\no MSE_u=\frac{1}{N_u}\sum_{i=1}^{N_u}\Big[~{\mid u(0,x_{i})-12\mathrm{sech}^{2}( x_{i})\mid}^{2}\\
&&\no\hspace{3.2cm}+{\mid u(t_{i},0)-12 \mathrm{sech}^{2} (-4t_{i})\mid}^{2}+{\mid u(t_{i},1)-12 \mathrm{sech}^{2} (1-4t_{i})\mid}^{2}~\Big],\\
&&\no MSE_f=\frac{1}{\widetilde{N}}\sum_{j=1}^{\widetilde{N}}{\mid f(\widetilde{t}_{j},\widetilde{x}_{j})\mid}^{2},~~~ MSE_g=\frac{1}{\widetilde{N}}\sum_{j=1}^{\widetilde{N}}{\mid g(\widetilde{t}_{j},\widetilde{x}_{j})\mid}^{2}.
\end{eqnarray}
where $MSE_u$ corresponds to the loss on the initial and boundary data $\{t_{i},x_{i},u^i\}_{i=0}^{N_u}$, $MSE_f$ penalizes the KdV equation not being satisfied on the collocation points $\{\widetilde{t}_{j},\widetilde{x}_{j}\}_{j=0}^{\widetilde{N}}$ while $MSE_g$ corresponds to the loss of the ISC on the same collocation points. Note that the loss function of PINN is $MSE=MSE_u+MSE_f$ which differentiates the one of SPINN with the ISC term $MSE_g$.  To obtain the training data set, we divide the spatial region $x\in[0, 1]$ and time region $t\in[0,1]$ into $N_x=256$ and $N_t=100$ discrete equidistance points respectively. Thus the solutions $u$ is discretized into $256 \times 100$ data points in the given spatio-temporal domain $[0,1]\times[0,1]$.

To investigate the performances of PINN and SPINN for the KdV equation \eqref{kdv}, we perform two groups of ten independent experiments to compare the prediction accuracy of the two methods where each initial seed is selected randomly, the training points $N_u=100$ are randomly sampled from the initial-boundary data set and the number of hidden layers is 2, and

Group I. The number of  collocation points $\widetilde{N}$ varies from 50 to 2050 with step 100 and each layer has 20 neurons;

Group II. The number of neurons in each layer changes from 10 to 100 with step 5 simultaneously and the  number of collocation points keeps $\widetilde{N}=700$ via the Latin hypercube sampling method \cite{ms-1987}.
\begin{figure}[htp]
	\begin{minipage}{0.5\linewidth}
		\vspace{3pt}
		\centerline{\includegraphics[width=\textwidth]{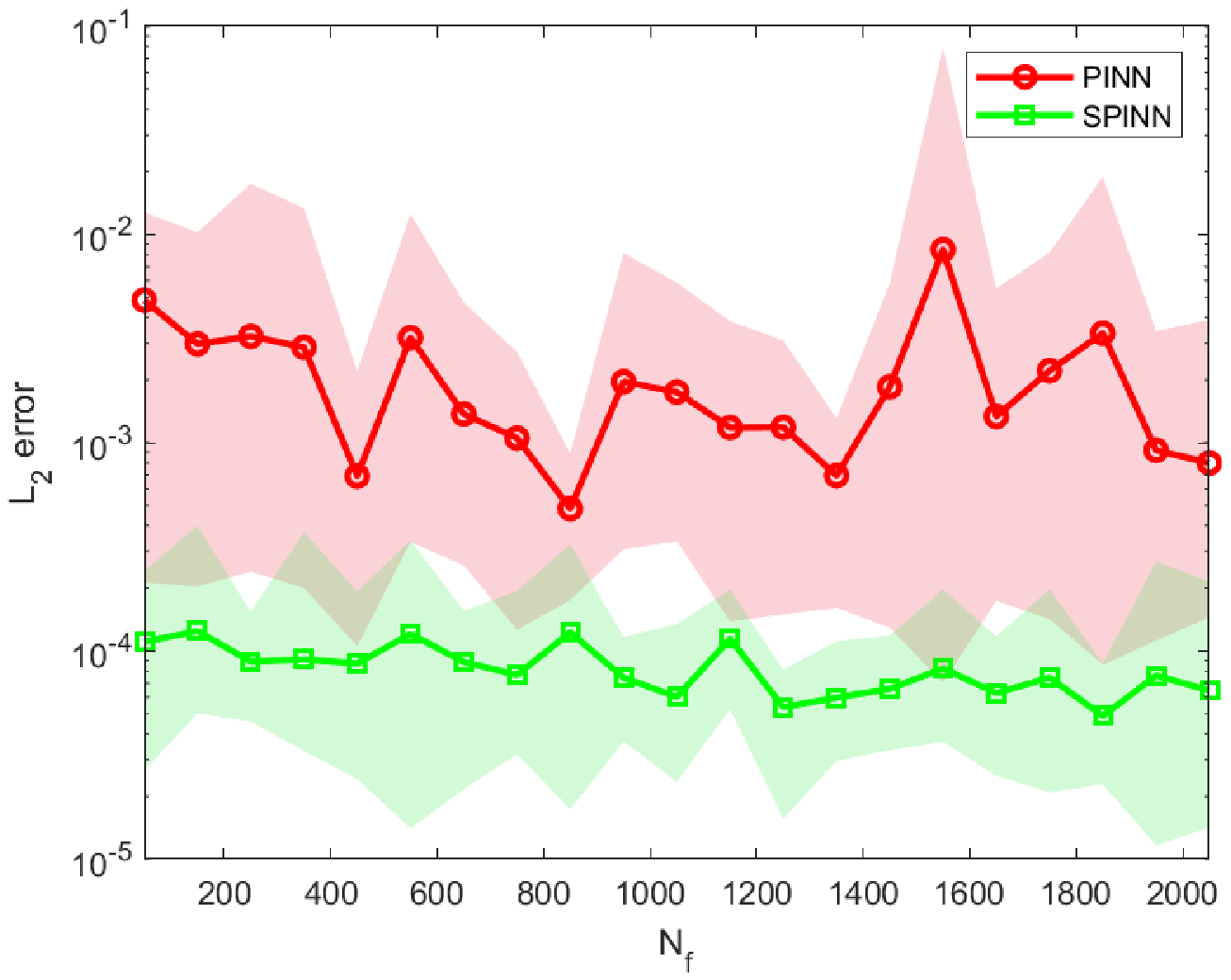}}
        \centerline{A}
	\end{minipage}
	\begin{minipage}{0.5\linewidth}
		\vspace{3pt}
		\centerline{\includegraphics[width=\textwidth]{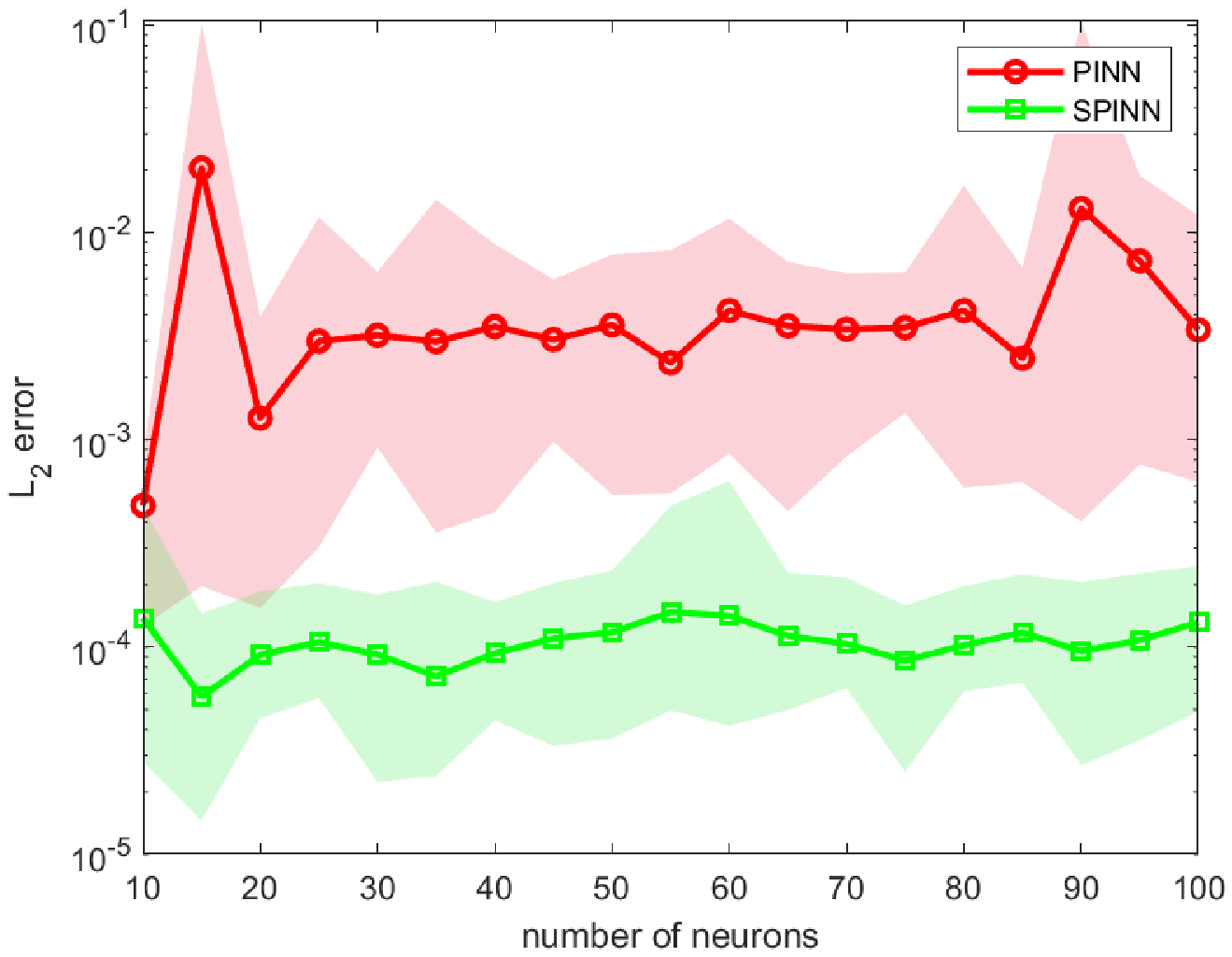}}
        \centerline{B}
	\end{minipage}
	\caption{(Color online) KdV equation: Comparison of $L_2$ relative errors of PINN and SPINN. (A) Keeping the number of {neurons} unchanged, the $L_2$ relative error of $u$ for PINN and SPINN using different numbers of collocation points. (B) Keeping the number of training points unchanged, the $L_2$ relative error of $u$ for PINN and SPINN using different numbers of neurons. The line and shaded region represent errors of the mean values and max-min values of 10 independent runs \cite{J}. }
\label{fig1-kdv}
\end{figure}

The $L_2$ relative errors of PINN and SPINN are displayed in Figure \ref{fig1-kdv} where the left one corresponds to Group I and the right one is for Group II. The red and green {lines} respectively correspond to the mean errors of ten experiments while the shade regions depict the max-min errors of ten experiments where the nodes show the locations of the collocation points or neurons for training. In Group I, the mean values of $L_2$ relative errors for SPINN outperform at least one order of magnitude than PINN, even to three orders at $\widetilde{N}=1500$. SPINN can reach $2.65\times10^{-5}$ $L_2$ relative error by using only 50 collocation points while PINN can not get the same accuracy within 2050 collocation points. Moreover, the fluctuation range of $L_2$ relative errors via the ten experiments for SPINN is much more smaller than PINN, which demonstrates that the $L_2$ relative error of SPINN is more stable than PINN. In Group II, as the increasing numbers of the neurons in graph B of Figure \ref{fig1-kdv}, the mean value of $L_2$ relative errors of SPINN always keep stable around $10^{-4}$ and far better than the ones of PINN which fluctuate around about $3\times10^{-3}$ and have big fluctuations at 10, 15 and 90 neurons respectively.

However, there exist intersections between the shade regions of SPINN and PINN, {because the best $L_2$ relative error of PINN and the worst one of SPINN just overlap in certain cases of ten experiments}. In fact, under the same conditions, SPINN still performs better than PINN. We illustrate it by selecting the two cases corresponding to both serious intersection and the worse error for SPINN in the two groups respectively and list their results explicitly in Table \ref{tab-kdv},
\begin{table}[htp]
    \centering
    \renewcommand{\arraystretch}{1.2}
    \caption{KdV equation: $L_2$ relative errors of PINN and SPINN and ERR}
    \begin{tabular}{l|lll}
        \hline
        \diagbox{Solution}{Method}& PINN  & SPINN & ERR\\
       \hline
        $u(\widetilde{N}=850$) & 3.704e-04 & 1.065e-04 & 71.24\% \\
        $u(10~ neurons$) & 1.100e-03 & 4.7646e-04 & 52.54\% \\
       \hline
    \end{tabular}
    \label{tab-kdv}
 \end{table}
where the error reduction rate (ERR) is computed according to the $L_2$ relative error of PINN ($Re_1$) and the one of SPINN ($Re_2$) \cite{jan},
\begin{eqnarray} \label{lieback1}
&& ERR=\frac{Re_1-Re_2}{Re_1}.
\end{eqnarray}

\begin{figure}[htp]
\centering
	\begin{minipage}{0.3\linewidth}
		\vspace{3pt}
		\centerline{\includegraphics[width=\textwidth]{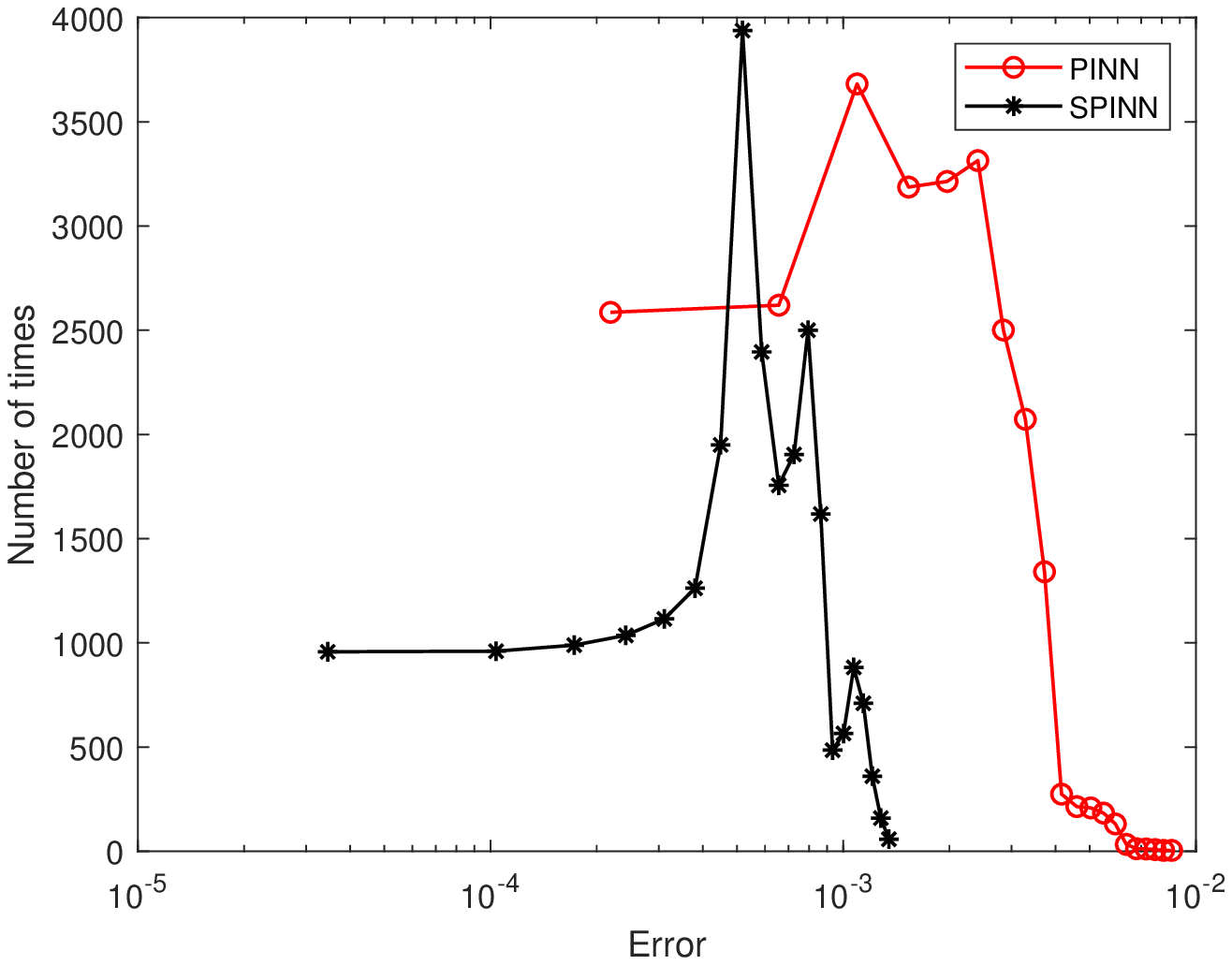}}
		\centerline{A}
	\end{minipage}
\begin{minipage}{0.3\linewidth}
		\vspace{3pt}
		\centerline{\includegraphics[width=\textwidth]{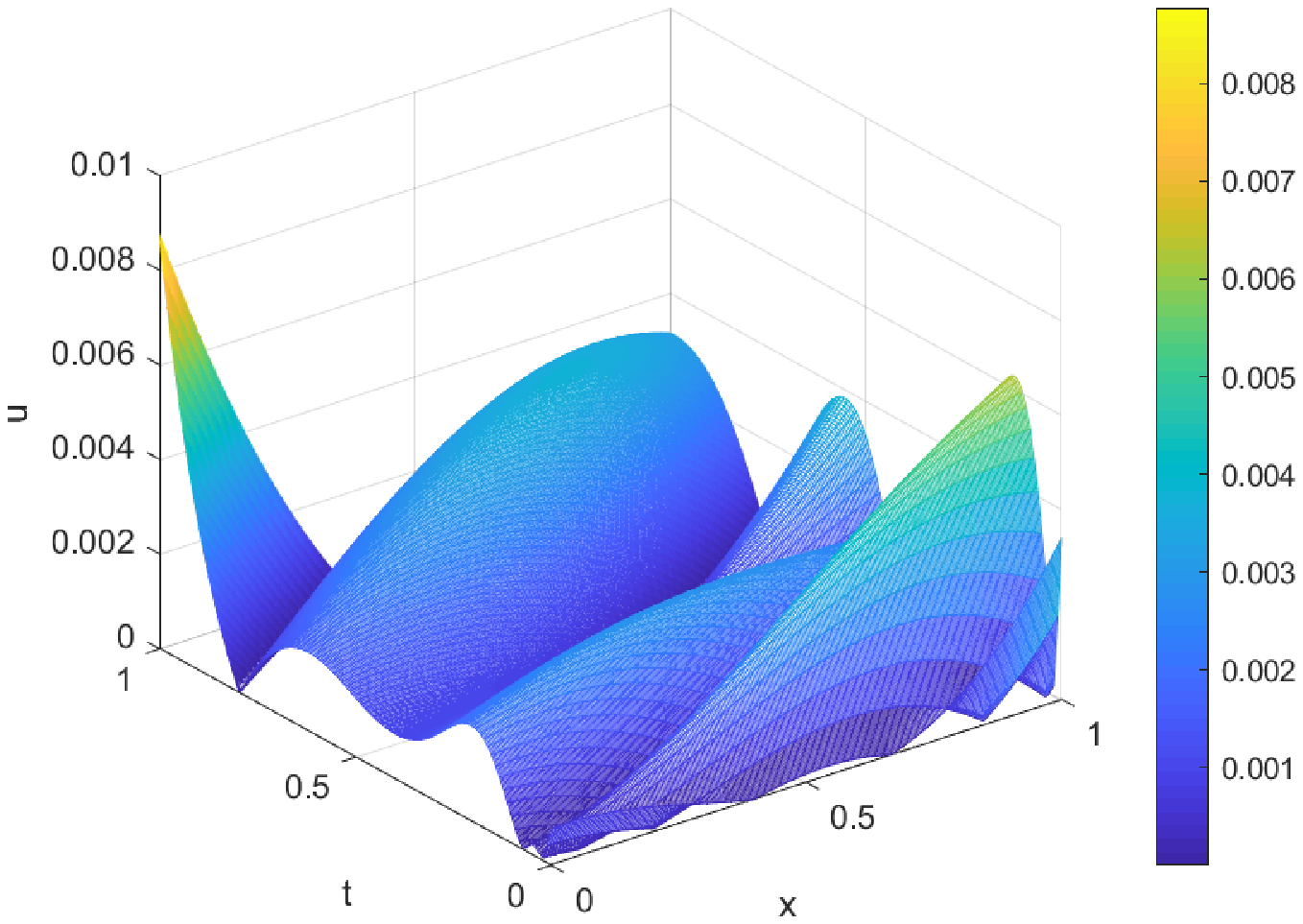}}
		\centerline{B: PINN}
	\end{minipage}
	\begin{minipage}{0.3\linewidth}
		\vspace{3pt}
		\centerline{\includegraphics[width=\textwidth]{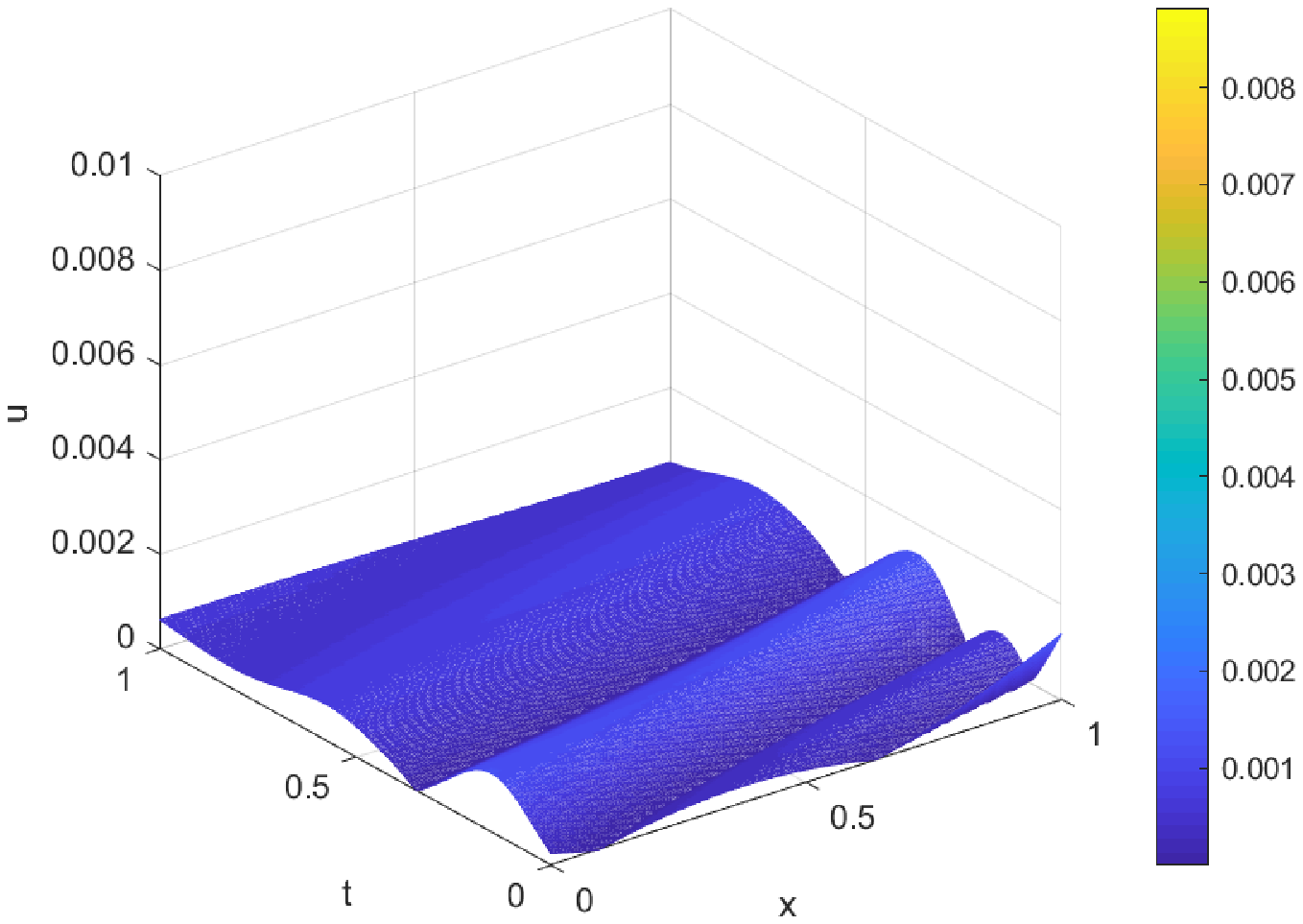}}
		\centerline{C: SPINN}
	\end{minipage}
\begin{minipage}{0.3\linewidth}
		\vspace{3pt}
		\centerline{\includegraphics[width=\textwidth]{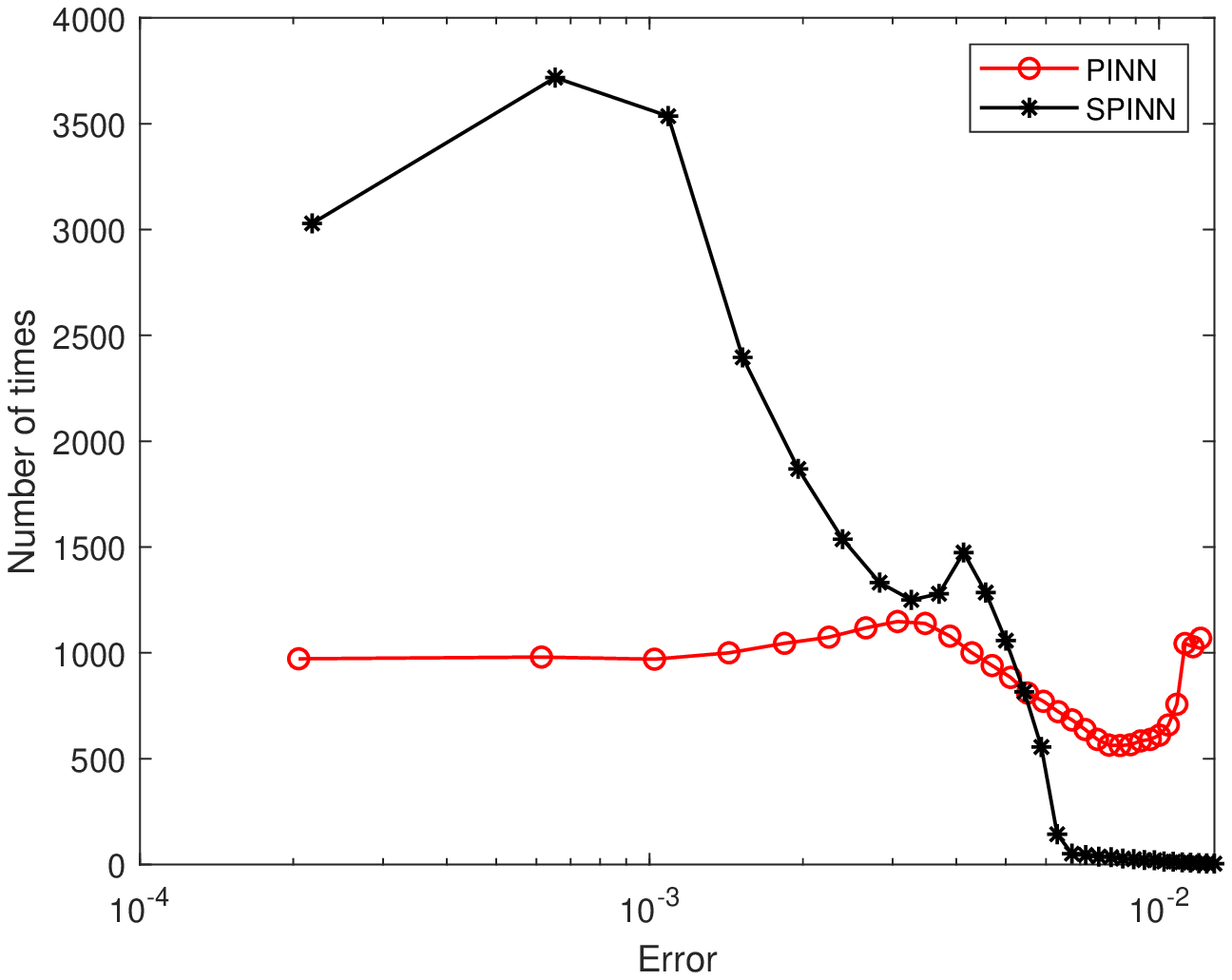}}
		\centerline{D}
	\end{minipage}
\begin{minipage}{0.3\linewidth}
		\vspace{3pt}
		\centerline{\includegraphics[width=\textwidth]{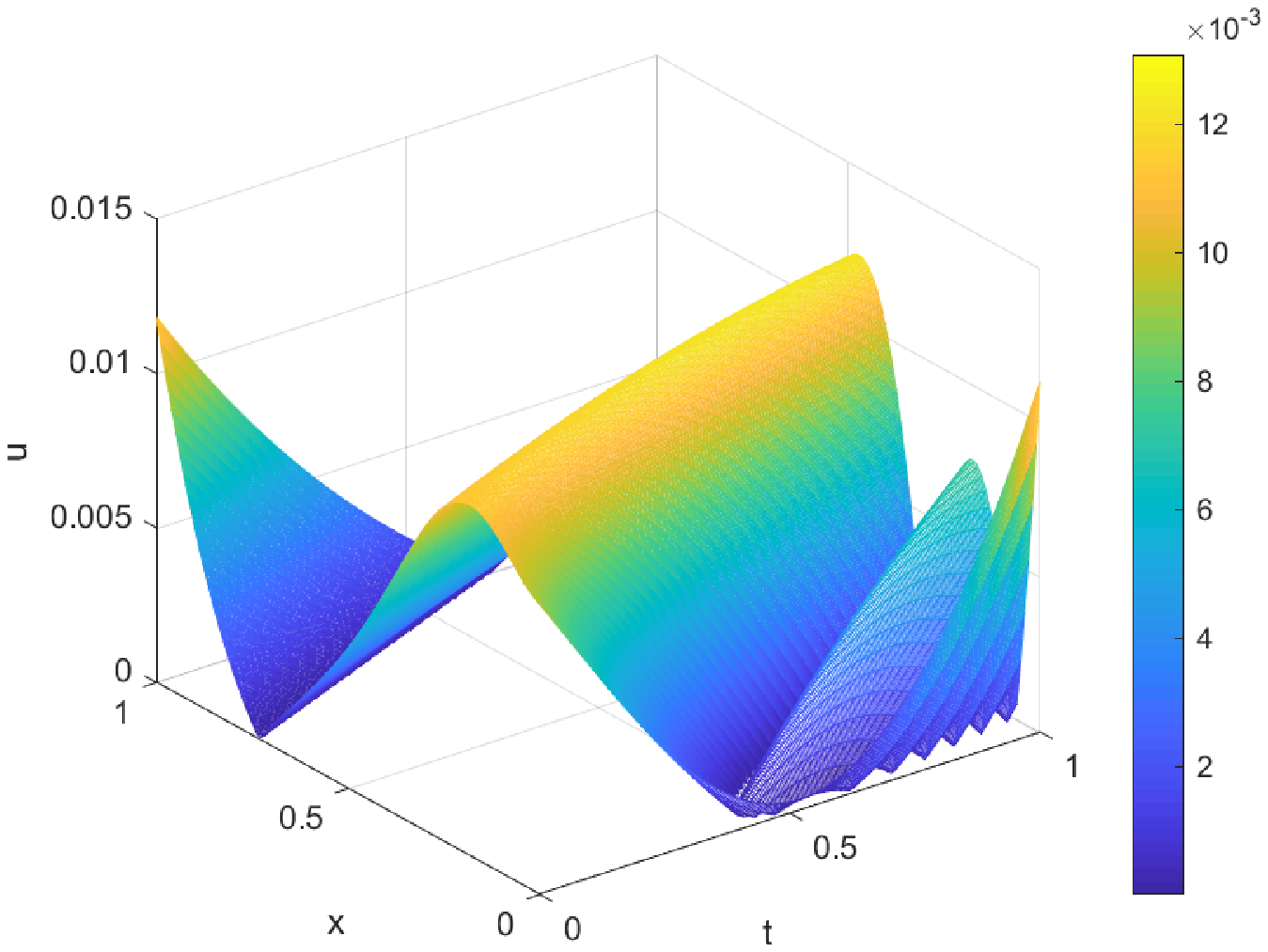}}
		\centerline{E: PINN}
	\end{minipage}
	\begin{minipage}{0.3\linewidth}
		\vspace{3pt}
		\centerline{\includegraphics[width=\textwidth]{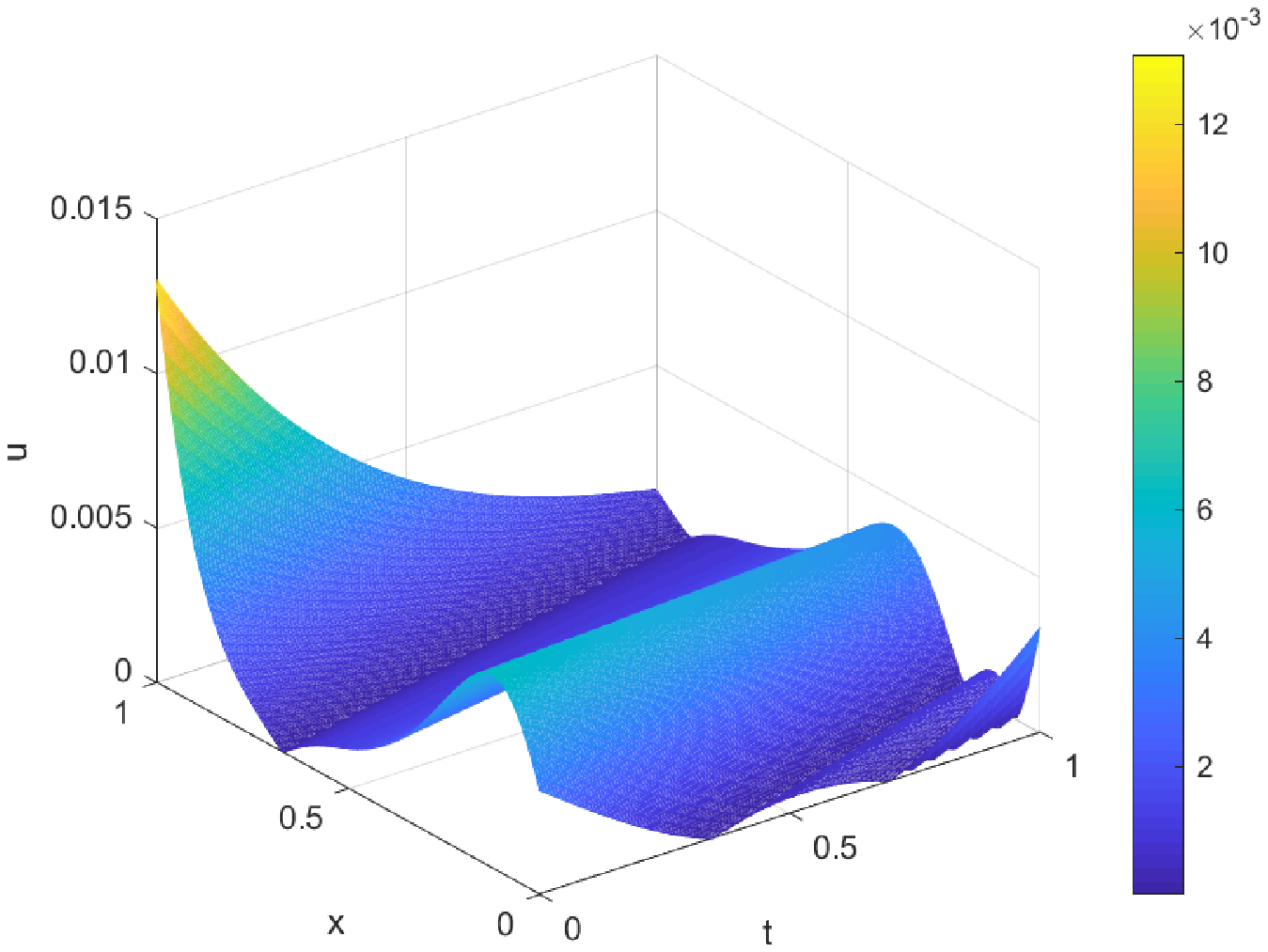}}
		\centerline{F: SPINN}
	\end{minipage}
	\caption{(Color online) KdV equation: Graph set ({A,B,C}) corresponds to the bad performance of SPINN  in Group I and ({D,E, F}) shows the one in Group II; (A and D): Absolute error distributions of learned solution $u$, ({B and C for Group I, E and F for Group II}) comparisons of absolute errors between PINN and SPINN.}
\label{fig2-kdv}
\end{figure}

Table \ref{tab-kdv} shows that, even in the two bad performances of SPINN, the ERR still has large drops, $71.24\%$ reduction at 850 collocation points and one order of magnitude reduction at 10 neurons. Furthermore, in the cases of two bad performances of SPINN, we compute the absolute errors of PINN and SPINN and show them in Figure \ref{fig2-kdv}. From the error distribution graph A for Group I in Figure \ref{fig2-kdv}, the absolute errors of the PINN  mainly distribute on the interval $(2.19\times10^{-4},8.54\times10^{-3})$ while the absolute errors of the SPINN gather on $(3.46\times10^{-5},1.35\times10^{-3})$. The peak value of SPINN appears at $5.18\times10^{-4}$ and the peak value of PINN emerges at $1.09\times10^{-3}$, {thus SPINN is more easily to get high accuracy numerical solutions of the KdV equation in view of collocation point. And the absolute errors of SPINN with one order of magnitude improvement appear much usually}.
For the Group II, in graph D, both the distributions of absolute errors for SPINN and PINN mostly gather at between negative four and negative two powers of ten, but the peak value of SPINN is $3718$ times  and corresponds to $6.53\times10^{-4}$ which is more close to the zero than the one of PINN, 1148 times corresponding to $3.07\times10^{-3}$,  thus SPINN has big advantages to find higher accuracy solutions of the KdV equation in terms of neurons. The three-dimensional distribution of absolute error in C (Group I) and F  (Group II) for SPINN are stable and all far better than the ones in  B (Group I) and E  (Group II) for PINN which have big fluctuations.

The experiment results for KdV equation show that enforcing the inherent physical properties, the ISC induced by the Lie symmetry, to the loss function can further constrain the MSE of PDEs and thus learn more accurate data-efficient solutions. As expected, the accuracy of the SPINN is higher than that of the PINN, and in some cases even three orders of magnitude higher. In addition, the travelling transformation, $\partial t+c\,\partial x$ with wave speed $c>0$, {leaves invariant all the PDEs in which the independent variables $t$ and $x$ do not appear explicitly,} thus the proposed SPINN has a good prospect of application and extension.
%
\subsection{Heat equation}
We consider the linear heat equation
\begin{eqnarray} \label{heat}
&& f:= u_t-u_{xx}=0,
\end{eqnarray}
subject to the initial and boundary conditions
\begin{eqnarray}\label{ib-heat}
&&\no u(\frac{1}{2},x)=2\sqrt{2}\,x\,\mathrm{e}^{-\frac{x^2}{2}},\\
&& u(t,0)=0,~~~ u(t,1)=t^{-\frac{3}{2}}\mathrm{e}^{-\frac{1}{4t}}.
\end{eqnarray}

Physically, the initial and boundary problems \eqref{heat} and \eqref{ib-heat} represent a model for the heat flow in an insulated wire of which the two ends are kept at $0^{\circ}$C and the time-dependent temperature $t^{-3/2}\mathrm{e}^{-1/(4t)}$ respectively, and the initial temperature distribution at $t=1/2$ is given as $2\sqrt{2}\,x\,\mathrm{e}^{-x^2/2}$ \cite{dv-1975}.

Eq.\eqref{heat} admits a Lie symmetry \cite{olv}
\begin{eqnarray}
&& \mathcal {X}_{heat}= xt\partial_x+ t^{2} \partial_t-\left(\frac{x^{2}}{4}+\frac{t}{2}\right)u\partial_u,
\end{eqnarray}
which generates an exact solution $u=xt^{-3/2}\mathrm{e}^{-\frac{x^2}{4t}}$. Meanwhile, the ISC induced by $\mathcal {X}_{heat}$ is
\begin{eqnarray} \label{hea}
&& g:=\left(\frac{x^{2}}{4}+\frac{t}{2}\right)u+xtu_x+t^2u_t=0.
\end{eqnarray}

To obtain the training data, we divide the time region $t\in[0.5,1.5]$ into $N_t=100$ and spatial region $x\in [0, 1]$ into $N_x=256$ equidistance points respectively, then the solution $u$ is discretized into $100\times256$ data points in the given spatio-temporal domain $[0.5,1.5]\times[0, 1]$.
The loss function of SPINN is given by
\begin{eqnarray} \label{inva}
&&\no MSE=MSE_u+MSE_f+MSE_g.
\end{eqnarray}
where
\begin{eqnarray}
&&\no MSE_u=\frac{1}{N_u}\sum_{i=1}^{N_u}\Big[~{\mid u(\frac{1}{2},x_{i})-2\sqrt{2}\,x_{i}\,\mathrm{e}^{-\frac{x_{i}^2}{2}}\mid}^{2}\\
&&\no\hspace{3.2cm}+{\mid u(t_{i},0)\mid}^{2}+{\mid u(t_{i},1)-t_{i}^{-\frac{3}{2}}\mathrm{e}^{-\frac{1}{4t_{i}}}\mid}^{2}~\Big],\\
&&\no MSE_f=\frac{1}{\widetilde{N}}\sum_{j=1}^{\widetilde{N}}{\mid f(\widetilde{t}_{j},\widetilde{x}_{j})\mid}^{2},~~~ MSE_g=\frac{1}{\widetilde{N}}\sum_{j=1}^{\widetilde{N}}{\mid g(\widetilde{t}_{j},\widetilde{x}_{j})\mid}^{2}.
\end{eqnarray}
where $MSE_u$ corresponds to the loss on the initial and boundary data $\{t_{i},x_{i},u^i\}_{i=0}^{N_u}$, $MSE_f$ and $MSE_g$ corresponds to the loss functions of  Eq.\eqref{heat} and the ISC on the collocation points $\{\widetilde{t}_{j},\widetilde{x}_{j}\}_{j=0}^{\widetilde{N}}$ respectively.

\begin{figure}[htp]
	\begin{minipage}{0.5\linewidth}
		\vspace{3pt}
		\centerline{\includegraphics[width=\textwidth]{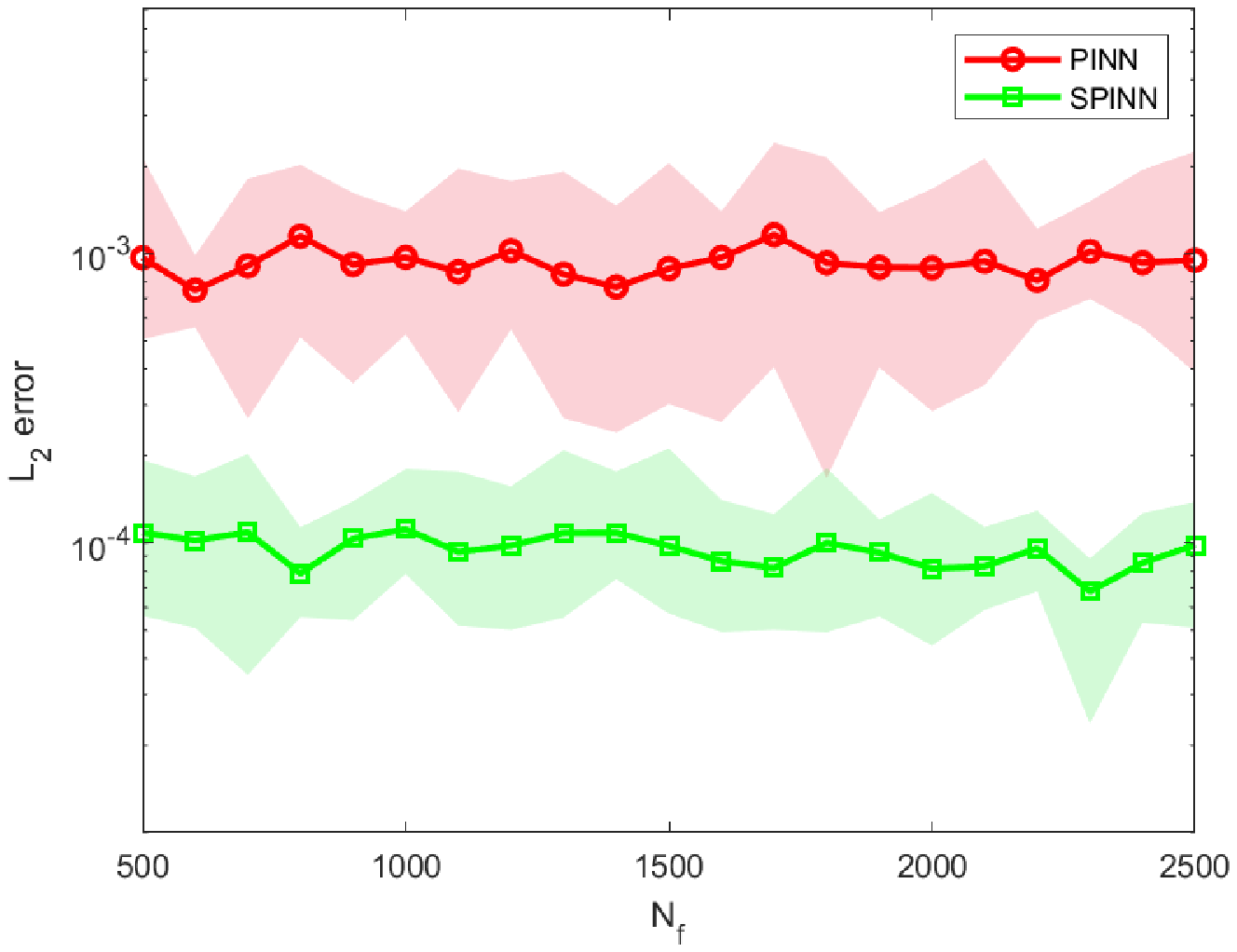}}
        \centerline{A}
	\end{minipage}
	\begin{minipage}{0.5\linewidth}
		\vspace{3pt}
		\centerline{\includegraphics[width=\textwidth]{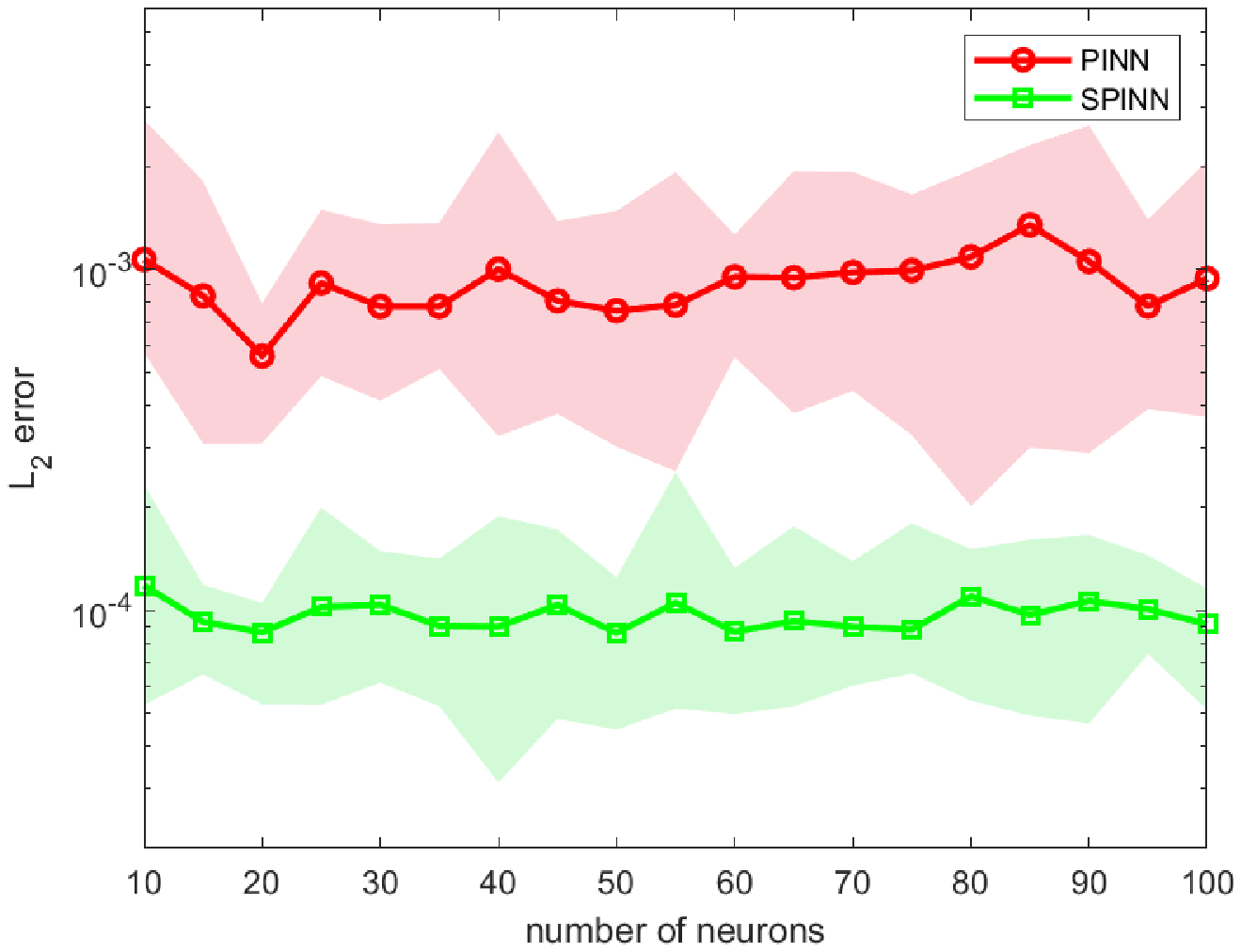}}
        \centerline{B}
	\end{minipage}
	\caption{(Color online) Heat equation: Comparison of $L_2$ relative errors of PINN and SPINN. (A) Keeping the number of training points unchanged, the $L_2$ relative errors of $u$ for PINN and SPINN using different numbers of collocation points. (B) Keeping the number of training points unchanged, the $L_2$ relative errors of $u$ for PINN and SPINN using different numbers of neurons. The line and shaded region represent the mean and max-min of 10 independent runs. }
\label{fig1-heat}
\end{figure}

To compare the performances of PINN and SPINN for Eq.\eqref{heat}, we perform two groups of ten independent experiments where one group aims at the influences of different numbers of collocation points, from 500 to 2500 with step 100, and the other considers the effects of the variation of number of neurons per layer, from 10 to 100 with step 5. In the experiments, the initial seeds are randomly selected to compare the $L_2$ relative errors of the two methods which are shown in Figure \ref{fig1-heat}.
In Figure \ref{fig1-heat}, the left graph shows the variations of $L_2$ relative errors of the two methods under different collocation points $\widetilde{N}$, where the training data $N_u=100$ are randomly sampled from the initial-boundary data set and the neural network is a 3-layer with 40 neurons per layer. The green line for SPINN slightly fluctuates around $10^{-4}$ which is particular better than the one of PINN which is depicted by red line with the mean value $10^{-3}$. The green shaded area denotes $L_2$ relative errors of ten experiments for SPINN and remains relatively steady, fluctuating from $2.38\times10^{-5}$ to $2.12\times10^{-4}$, while the amplitude of red shaded for PINN is obviously amplified. For the group of variations of neurons, we randomly select $N_u=150$ training points and proceed by sampling $\widetilde{N}=1000$ collocation points via the Latin hypercube sampling method, and fix the deep neural network as 3-layer. The right graph in Figure \ref{fig1-heat}  shows the effects of variations of neurons for the two methods where the green line of SPINN have no big fluctuations and is far below the red line of PINN. Except for the unhoped case $\widetilde{N}=1800$ in graph A, the two shade regions have no intersection totally which demonstrate that SPINN outperform PINN better.

\begin{figure}[htp]
\centering
	\begin{minipage}{0.3\linewidth}
		\vspace{3pt}
		\centerline{\includegraphics[width=\textwidth]{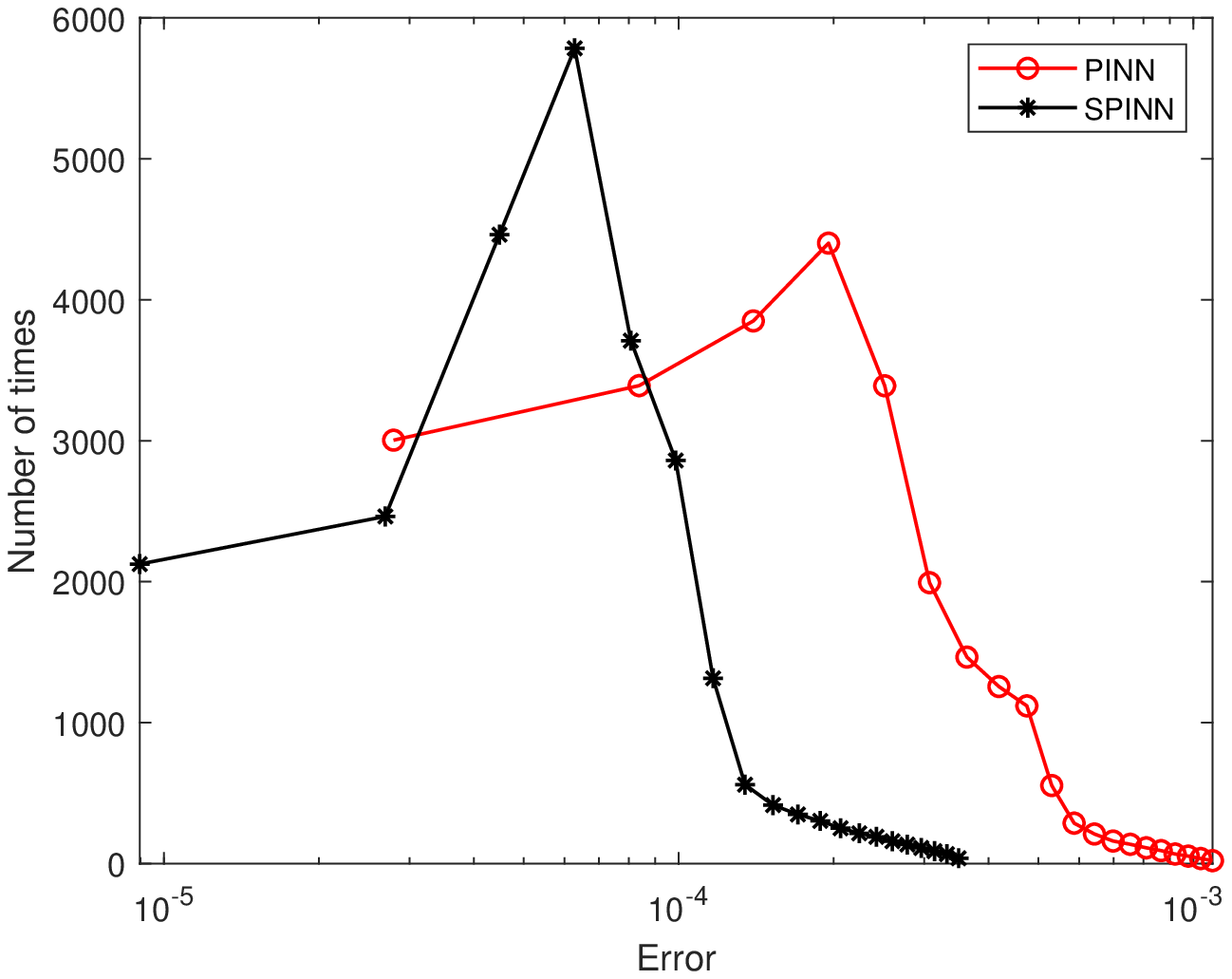}}
		\centerline{A}
	\end{minipage}
\begin{minipage}{0.33\linewidth}
		\vspace{3pt}
		\centerline{\includegraphics[width=\textwidth]{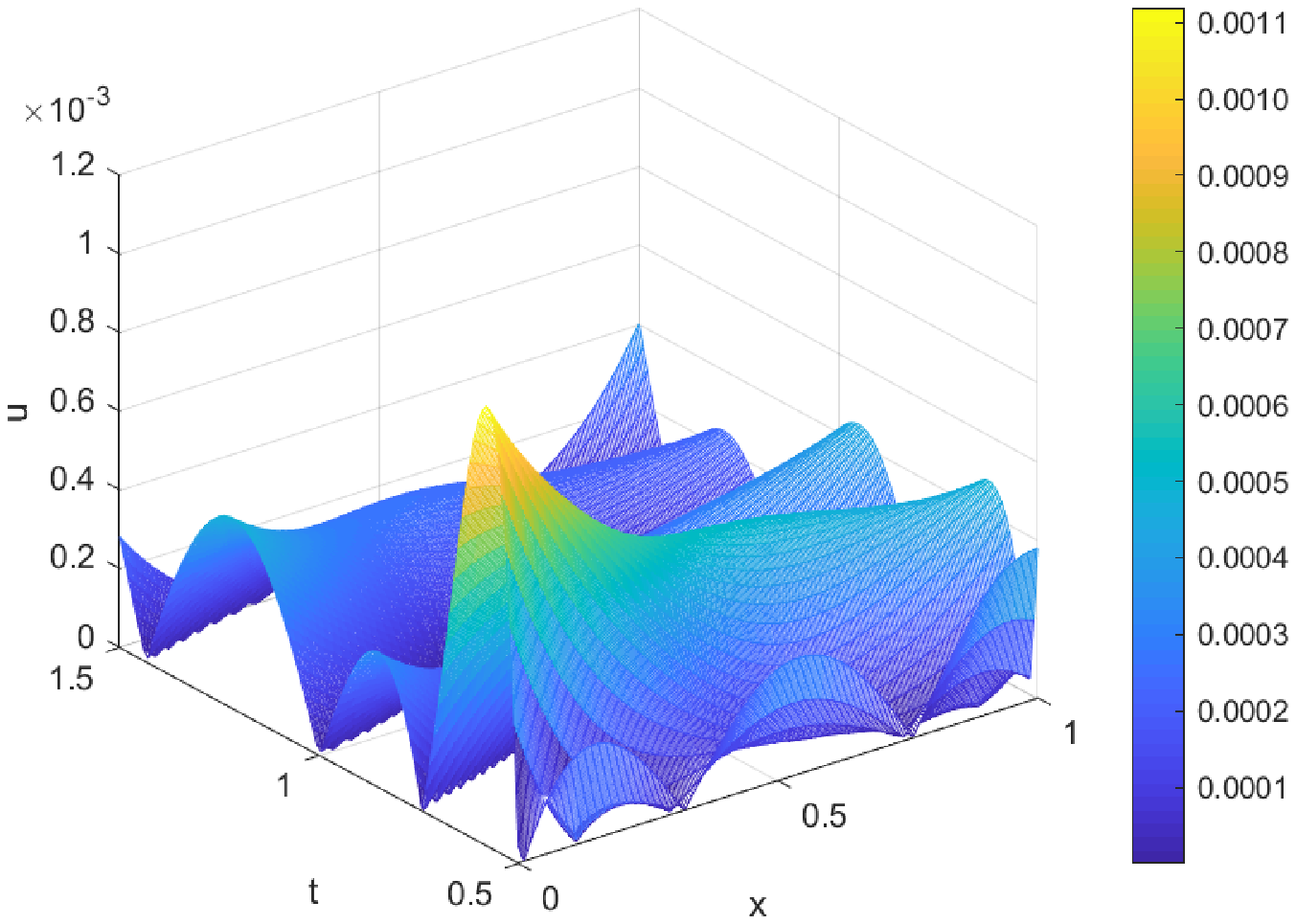}}
		\centerline{B: PINN}
	\end{minipage}
	\begin{minipage}{0.33\linewidth}
		\vspace{3pt}
		\centerline{\includegraphics[width=\textwidth]{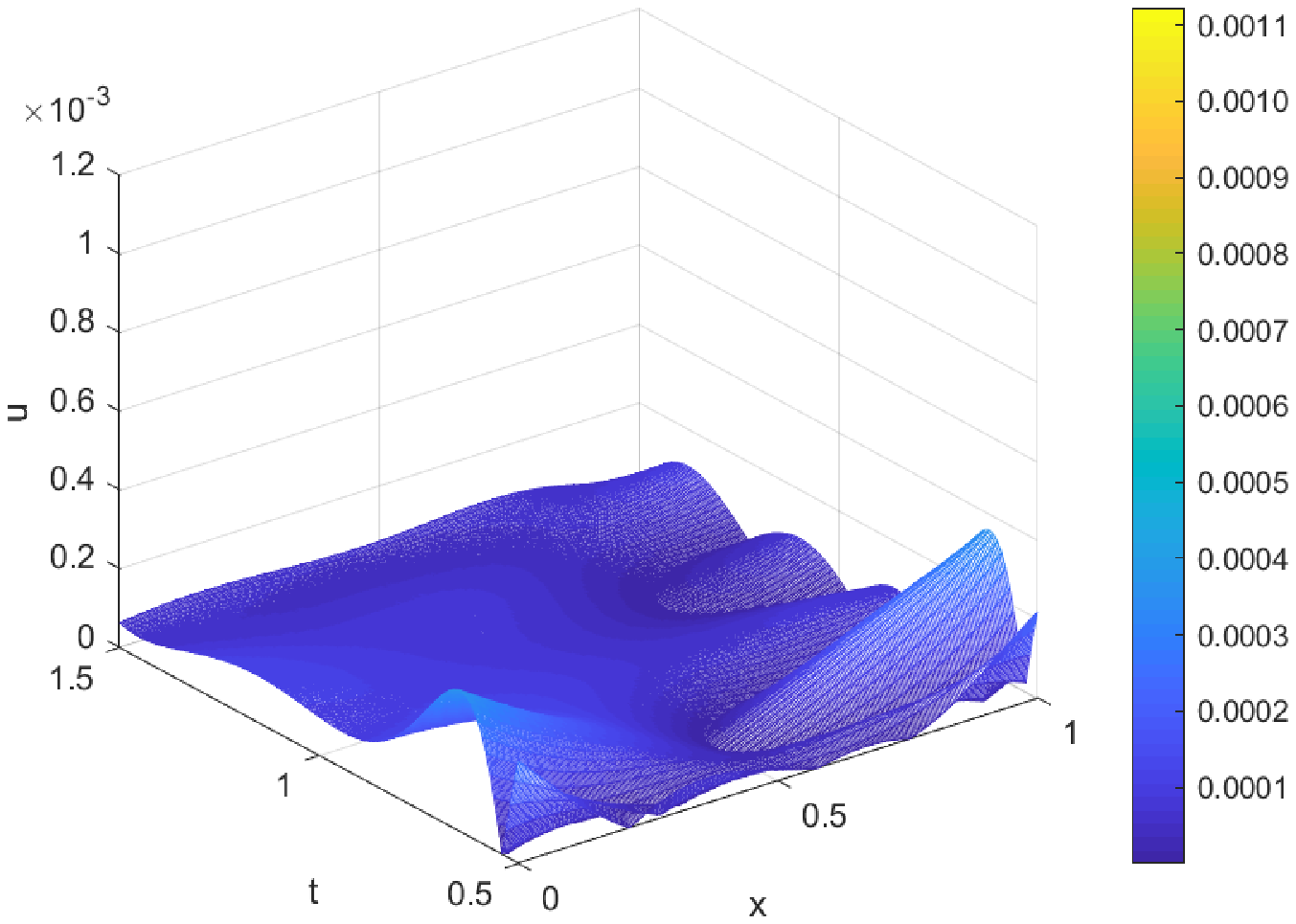}}
		\centerline{C: SPINN}
	\end{minipage}
	\caption{(Color online) Heat equation: The graph set ({A,B,C}) correspond to the selected bad performances of SPINN in the left experiment in Figure \ref{fig1-heat}. (A): Absolute error distributions of $u$ for the bad performance. Comparisons of absolute errors between PINN and SPINN for the bad performance of SPINN (B and C for the left experiments in Figure \ref{fig1-heat}).}
\label{fig1-heata}
\end{figure}

In the above two groups of experiments, it is easy to find that SPINN is much better than PINN in ten experiments under different network structures. However, ten experiments on collocation points obviously have intersection points, so we chose the serial intersection and the worse error for SPINN, and listed the $L_2$ relative error and ERR defined by (\ref{lieback1}) in Table \ref{tab-heat}.
\begin{table}[htp]
    \centering
    \renewcommand{\arraystretch}{1.2}
    \caption{The heat equation: $L_2$ relative errors of PINN and SPINN and ERR}
    \begin{tabular}{l|lll}
        \hline
        \diagbox{Solution}{Method}& PINN  & SPINN & ERR\\
       \hline
        $u(\widetilde{N}=1800$) & 4.5847e-04 & 1.4963e-04 & 67.36\% \\
       \hline
    \end{tabular}
    \label{tab-heat}
 \end{table}
Even in the bad performance of SPINN there has big improvements of $L_2$ relative error, i.e. $67.36\%$. Furthermore, {the absolute error} of the selected bad case of SPINN are depicted in Figure \ref{fig1-heata}. The graph A shows the absolute error distributions of the bad performances of SPINN, where the distribution of SPINN in the group is much more close to zero. It is obvious that SPINN has greater superiority than PINN. The graph B further shows the absolute error surface of $u$ for PINN, where it has big fluctuations and thus PINN is sensitive to the variations of collocation points, but the graph C for SPINN exhibit flat tendency. Obviously, the absolute error is more stable for the experiment of collocation point.


%
\subsection{Potential Burgers equations}
The last example is the nonlinear potential Burgers equations
\begin{eqnarray}\label{burgers}
&&\no f(t,x):=v_{x}-u=0, \\
&& g(t,x):=v_{t}-u_{x}+\frac{1}{2}u^{2}=0,
\end{eqnarray}
together with the initial and boundary conditions
\begin{eqnarray}\label{ib-burger}
&&\no u(0,x)=-\frac{4(x+2)}{x(4+x)}, ~v(0,x)=-2\ln\left(\frac{x}{3}+\frac{x^{2}}{12}\right), \\
&&\no u(t,0.1)= \frac{-840}{200t+41},~v(t,0.1)=-2\ln\left({\frac{t}{6}}+ {\frac{41}{1200}}\right),\\
&& u(t,1.1)={\frac{-1240}{200t+561}},~v(t,1.1)=-2\ln\left({\frac{t}{6}}+{\frac{187}{400}}\right).
\end{eqnarray}
The compatible condition $v_{xt}=v_{tx}$ in system \eqref{burgers} yields the celebrated Burgers equation $u_t+uu_x-u_{xx}=0$ which is the simplest wave equation combing both dissipative and nonlinear effects \cite{olv}, and also describes the interaction of convection and diffusion in turbulent fluid and is similar with Navier-Stokes equations \cite{burger-1948}.

System \eqref{burgers} admits a non-classical symmetry \cite{mm-2009}
$$ \mathcal {X}_{burgers}= -\frac{1}{x+1}\partial x+ \partial t+\frac{1}{(x+1)^{2}}\left[\frac{1}{6}(x+1)ue^{\frac{v}{2}}-u-\frac{1}{3}e^{\frac{v}{2}}\right]\partial u+\frac{1}{3(x+1)}e^{\frac{v}{2}}\partial v,$$
which generates a set of exact solution
\begin{equation}
 u(t,x)={\frac{-4(x+2)}{2t+4x+x^2}}, ~~~v(t,x)=-2\ln\left({\frac{t}{6}}+{\frac{x}{3}}+{\frac{x^{2}}{12}}\right).
\end{equation}
The ISCs associated with the non-classical symmetry $\mathcal {X}_{burgers}$ are
\begin{eqnarray}
&&\no l(t,x):= v_{t}-\frac{1}{x+1} v_{x}-\frac{1}{3(x+1)}e^{\frac{v}{2}}=0, \\
&&\no p(t,x):=u_{t}-\frac{1}{x+1} u_{x} -\frac{1}{(x+1)^{2}}\left[\frac{1}{6}(x+1)ue^{\frac{v}{2}}-u-\frac{1}{3}e^{\frac{v}{2}}\right]=0.
\end{eqnarray}

The shared parameters of the neural network $(u(t, x),v(t, x))$ can be learned by minimizing the mean squared error loss
\begin{eqnarray} \label{intr}
&& MSE=MSE_u+MSE_v+MSE_f+MSE_g+MSE_l+MSE_p,
\end{eqnarray}
where $MSE_u$ and $MSE_v$ correspond to the loss on the initial and boundary data $\{t_{i},x_{i},u^i,v^i\}_{i=0}^{\widehat{N}}$, $MSE_f$ and $MSE_g$ penalize the Burgers potential equations not being satisfied on the collocation points$\{\widetilde{t}_{j},\widetilde{x}_{j}\}_{j=0}^{\widetilde{N}}$, $MSE_l$ and $MSE_p$ correspond to the loss of ISC,
\begin{eqnarray}
&&\no MSE_u=\frac{1}{\widehat{N}}\sum_{i=1}^{\widehat{N}}\Big[~{\mid u(0,x_{i})-\frac{4(x_{i}+2)}{x_{i}(4+x_{i})}\mid}^{2}+{\mid u(t_{i},0.1)}+{\frac{840}{200t_{i}+41}\mid}^{2}\\
&&\no\hspace{3.2cm}+{\mid u(t_{i},1.1)+{\frac{1240}{200t_{i}+561}}\mid}^{2}~\Big],\\
&&\no MSE_v=\frac{1}{\widehat{N}}\sum_{i=1}^{\widehat{N}}\Big[~{\mid v(0,x_{i})+2\ln\left(\frac{x_{i}}{3}+\frac{x_{i}^{2}}{12}\right)\mid}^{2}+{\mid v(t_{i},0.1)+2\ln\left({\frac{t_{i}}{6}}+ {\frac{41}{1200}}\right)\mid}^{2}\\
&&\no\hspace{3.2cm}+{\mid v(t_{i},1.1)+2\ln\left({\frac{t_{i}}{6}}+{\frac{187}{400}}\right)\mid}^{2}~\Big],\\
&&\no MSE_f=\frac{1}{\widetilde{N}}\sum_{j=1}^{\widetilde{N}}{\mid f(\widetilde{t}_{j},\widetilde{x}_{j})\mid}^{2},~~~ MSE_g=\frac{1}{\widetilde{N}}\sum_{j=1}^{\widetilde{N}}{\mid g(\widetilde{t}_{j},\widetilde{x}_{j})\mid}^{2},\\
&&\no MSE_l=\frac{1}{\widetilde{N}}\sum_{j=1}^{\widetilde{N}}{\mid l(\widetilde{t}_{j},\widetilde{x}_{j})\mid}^{2},~~~ MSE_p=\frac{1}{\widetilde{N}}\sum_{j=1}^{\widetilde{N}}{\mid p(\widetilde{t}_{j},\widetilde{x}_{j})\mid}^{2}.
\end{eqnarray}

\begin{figure}[htp]
	\begin{minipage}{0.5\linewidth}
		\vspace{3pt}
		\centerline{\includegraphics[width=\textwidth]{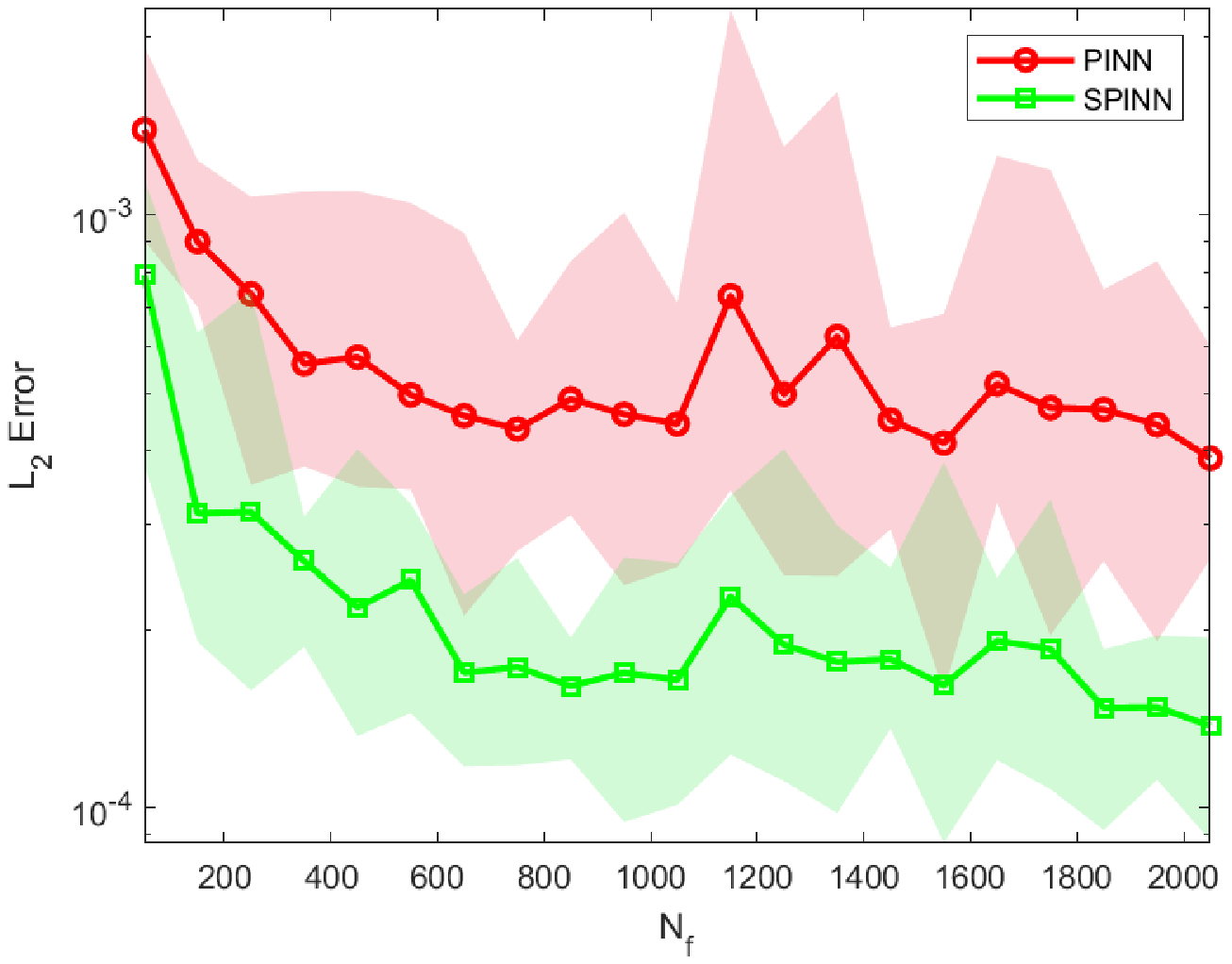}}
        \centerline{A}
	\end{minipage}
	\begin{minipage}{0.5\linewidth}
		\vspace{3pt}
		\centerline{\includegraphics[width=\textwidth]{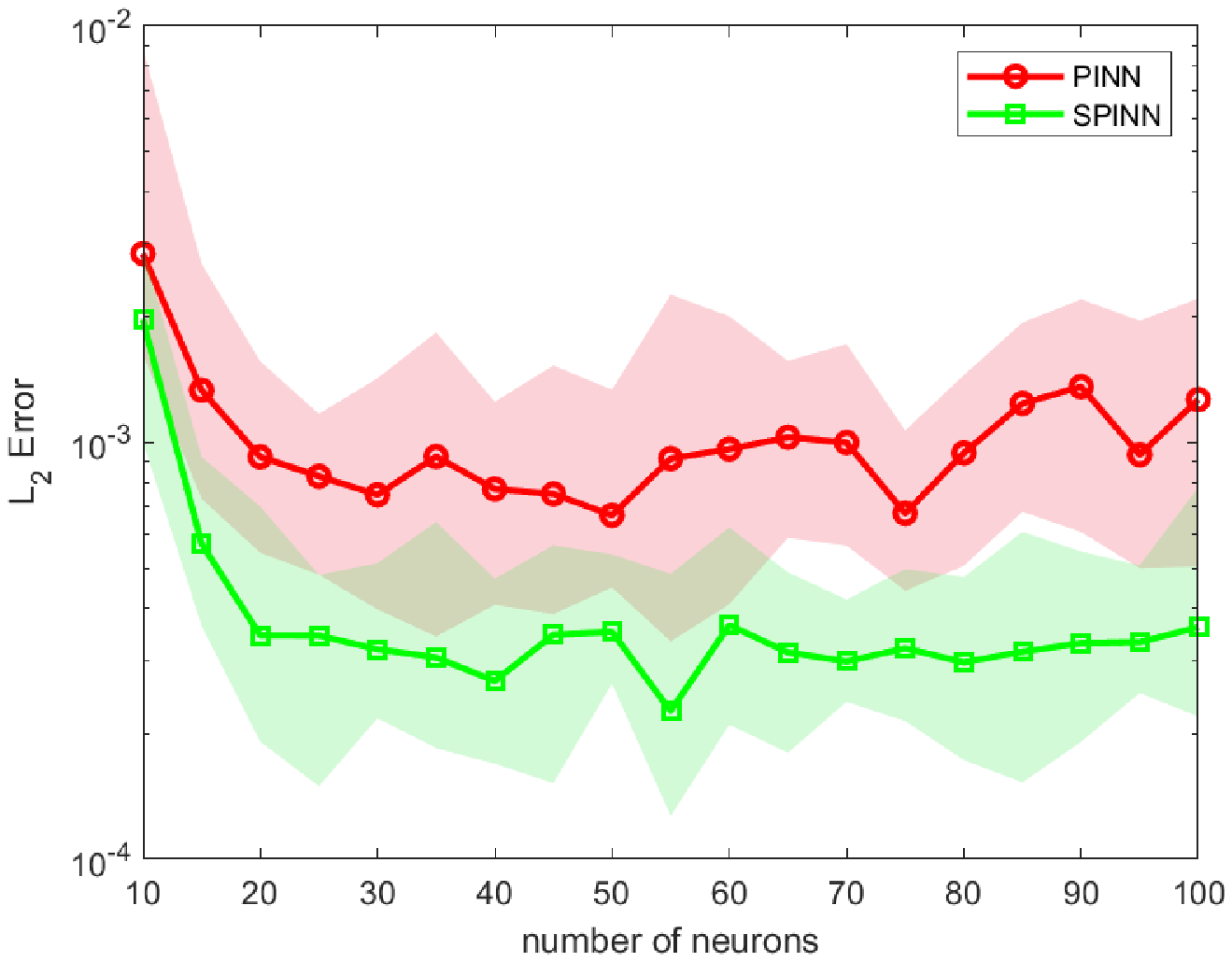}}
        \centerline{B}
	\end{minipage}
	\caption{(Color online) Potential Burgers equations: Comparison of the $L_2$ relative errors of PINN and SPINN. (A) Keeping the number of training points unchanged, the sum of the $L_2$ relative errors of $u$ and $v$ for PINN and SPINN using different numbers of collocation points. (B) Keeping the number of layers unchanged, the sum of the $L_2$ relative errors of $u$ and $v$ for PINN and SPINN using different numbers of neurons. The line and shaded region represent the mean and max-min of 10 independent runs. }
\label{fig1-burgers}
\end{figure}

To obtain the training data, we divide the spatial region $x\in[0.1, 1.1]$ and time region $t\in[0,1]$ into $N_x=256$ and $N_t=100$ discrete equidistance points respectively. Thus, the solutions $u$ and $v$ are discredited into $256\times100$ data points in the given spatio-temporal domain $[0.1, 1.1]\times[0,1]$.  In what follows, the variations of two indexes, collocation points and neurons, are used to verify the overall effectiveness of SPINN for system (\ref{burgers}).
We exhibit the $L_2$ relative error, defined by $error_u+error_v$ where $error_u$ and $error_v$ stand for $L_2$ relative error of $u$ and $v$ respectively,  via PINN and SPINN in the left graph in Figure \ref{fig1-burgers}, where the numbers of collocation points $\widetilde{N}$ vary from 50 to 2050 with step 100, the number of training points keeps unchanged $N_u=100$ randomly sampled from the initial and boundary data set, and the neural network architecture is 2 hidden layers with 60 neurons per layer. The green line for SPINN, the mean value of ten independent runs with random seeds, is far below the red line which stands for the mean value by PINN. The $L_2$ relative errors for SPINN witness a downward trend as the increasing of collocation points. Moreover, SPINN reaches $L_2$ relative error $1.32\times10^{-4}$ by using only 450 collocation points, the bottom of the green shadow at 450, while PINN can not reach the same accuracy with 2050 collocation points.

The right graph in Figure \ref{fig1-burgers} shows $L_2$ relative error for the 2-layer neural network with the number of neurons per layer, varying from 10 to 100 with step 5, where the randomly training points are $\widehat{N}=100$ and the collocation points are $\widetilde{N}=500$ by means of the Latin hypercube sampling method. Again, the green line for SPINN is totally below the red line for PINN, and SPINN gets $1.9\times10^{-4}$ by using only 20 neurons while PINN can not reach the same accuracy with 100 neurons. Moreover, as the numbers of neuron increase in graph B of Figure \ref{fig1-burgers}, the mean value of $L_2$ relative errors for SPINN  keeps stable after $60$ neurons but the one for PINN still has fluctuations. 

However, we cannot guarantee that, among the 10 experiments, the worst $L_2$ relative error of SPINN is better than the best one of PINN, which leads to the intersection of the two shaded areas in Figure \ref{fig1-burgers}. However, it is true that in the case of same conditions, the error of SPINN is still lower than that of PINN. We might as well respectively choose the cases corresponding to both serious intersection and the worst error of SPINN from the two sets of experiments, and take them out separately to compare the $L_2$ relative error and absolute error with PINN.
Table \ref{tab-burgers} shows the comparisons of SPINN with PINN for the two bad performances where the ERR is defined by (\ref{lieback1}) and the improvements of $L_2$ relative errors in both cases for $u$ is better than $v$.
\begin{table}[htp]
    \centering\renewcommand{\arraystretch}{1.2}
    \caption{Potential Burgers equations: $L_2$ relative errors of PINN and SPINN and ERR.}
    \begin{tabular}{l|lll}
        \hline
        \diagbox{Solution}{Method}& PINN  & SPINN &ERR\\
       \hline
       $ u(\widetilde{N}=250$) & 7.462e-04 & 1.954e-04 &  73.81\% \\
        $v(\widetilde{N}=250$) & 3.619e-04 & 2.099e-04 &   42.00\% \\
       $ u(neurons=35$) & 4.457e-04 & 9.613e-05  &78.43\% \\
        $v(neurons=35$) & 2.351e-04 & 1.050e-04 & 55.34\% \\
       \hline
    \end{tabular}
    \label{tab-burgers}
\end{table}

\begin{figure}[hpb]
\centering
	\begin{minipage}{0.3\linewidth}
		\vspace{3pt}
		\centerline{\includegraphics[width=\textwidth]{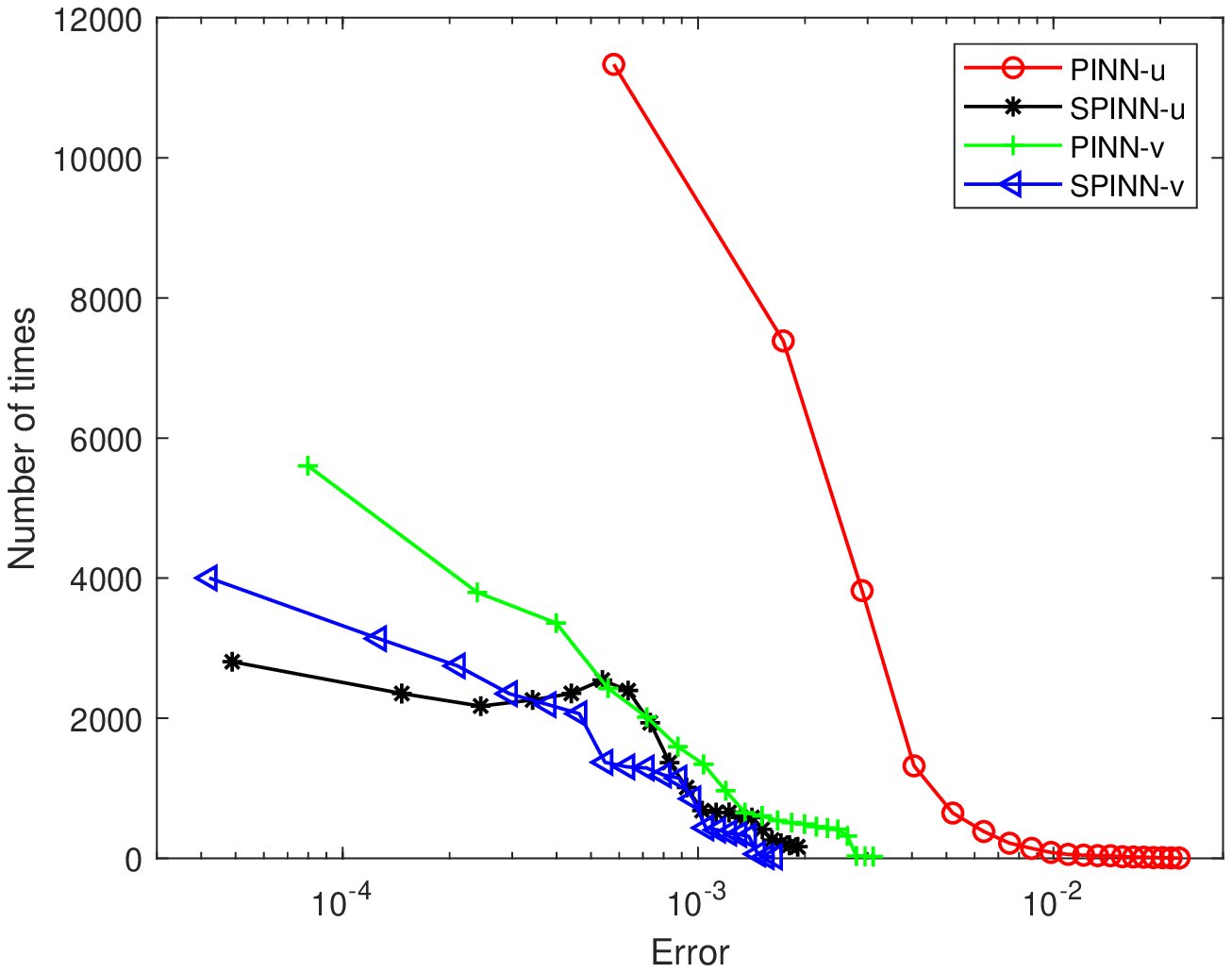}}
		\centerline{A}
	\end{minipage}
\begin{minipage}{0.33\linewidth}
		\vspace{3pt}
		\centerline{\includegraphics[width=\textwidth]{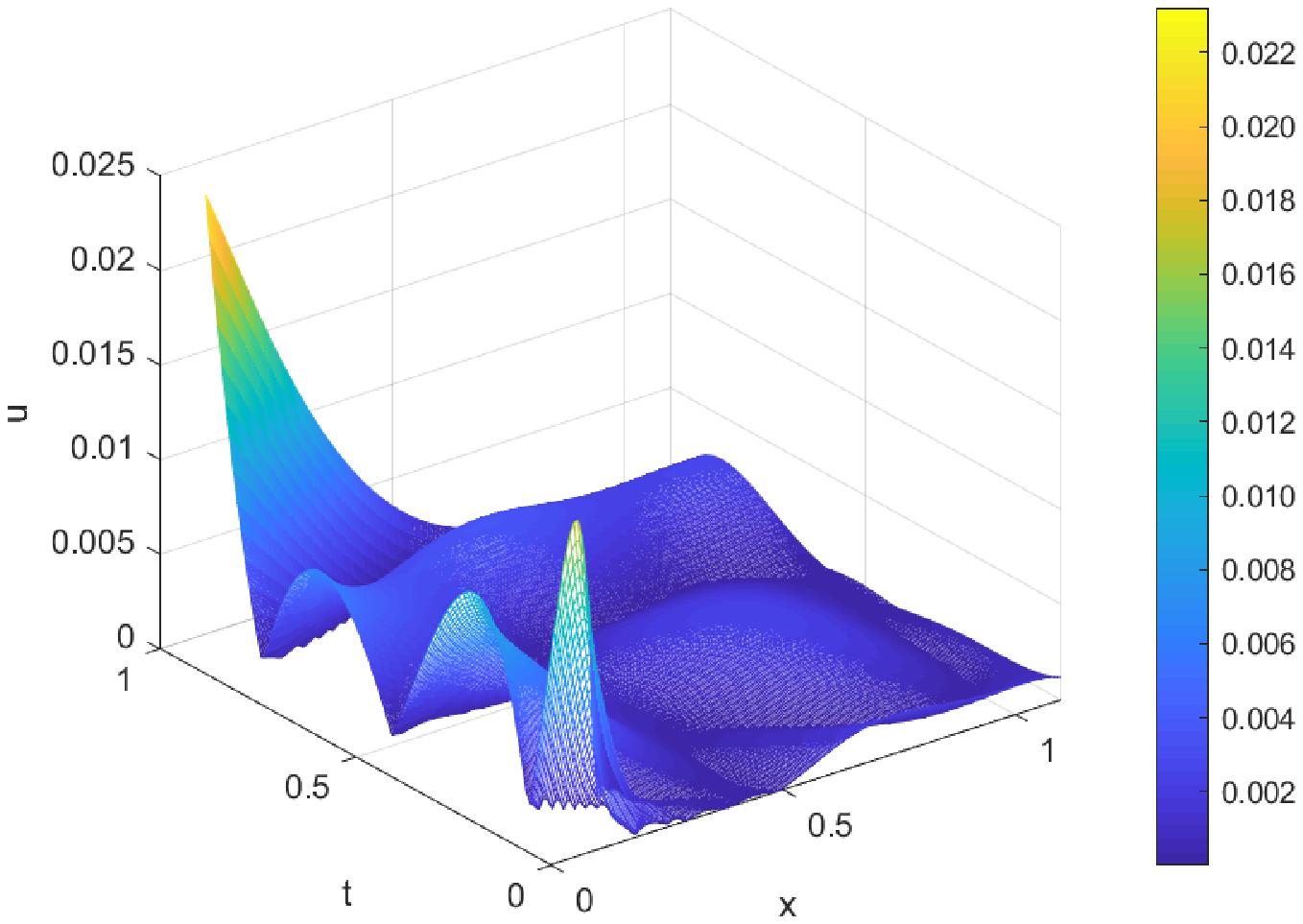}}
		\centerline{B: PINN$_u$}
	\end{minipage}
\begin{minipage}{0.33\linewidth}
		\vspace{3pt}
		\centerline{\includegraphics[width=\textwidth]{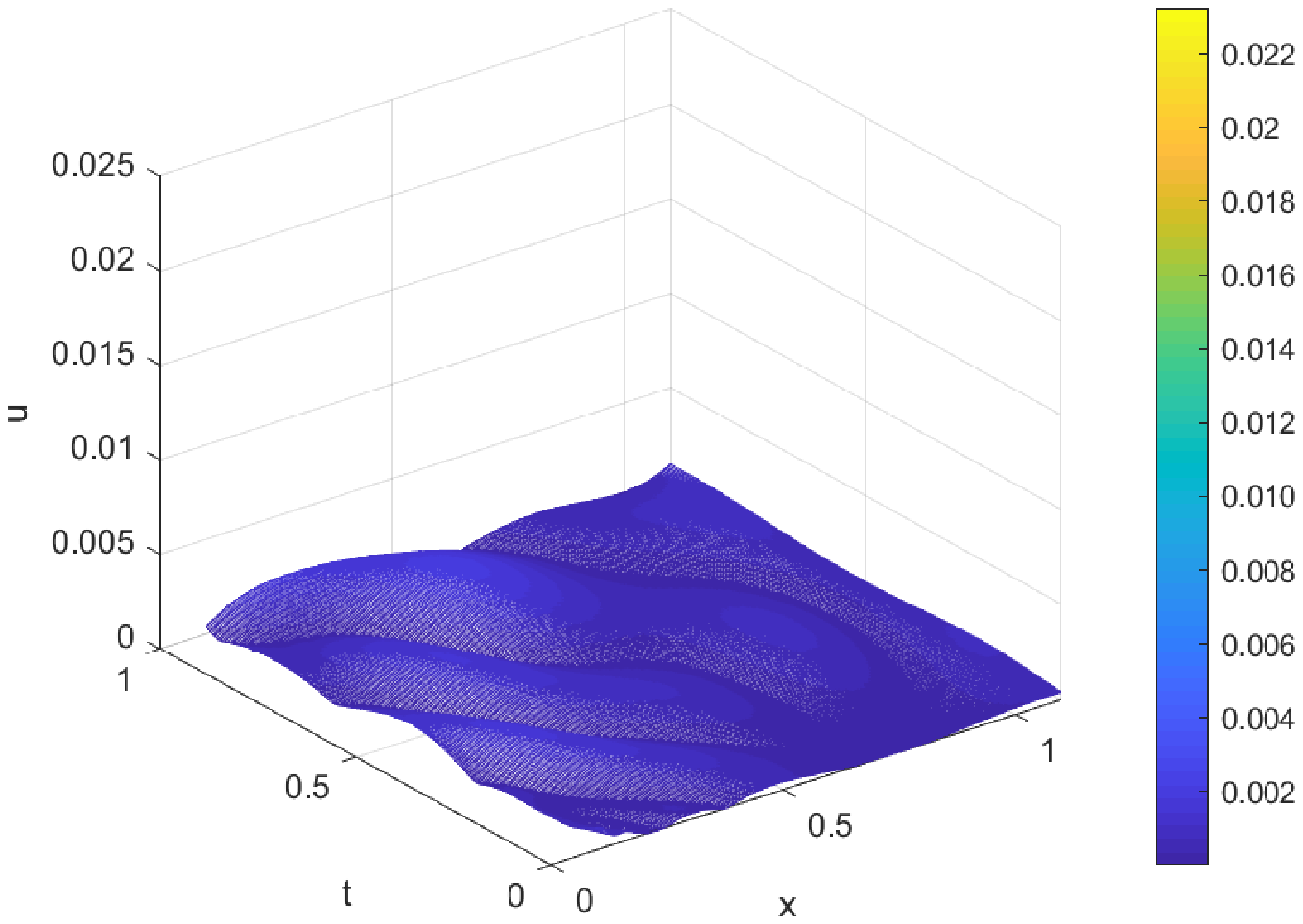}}
		\centerline{C: SPINN$_u$}
	\end{minipage}
	\begin{minipage}{0.33\linewidth}
		\vspace{3pt}
		\centerline{\includegraphics[width=\textwidth]{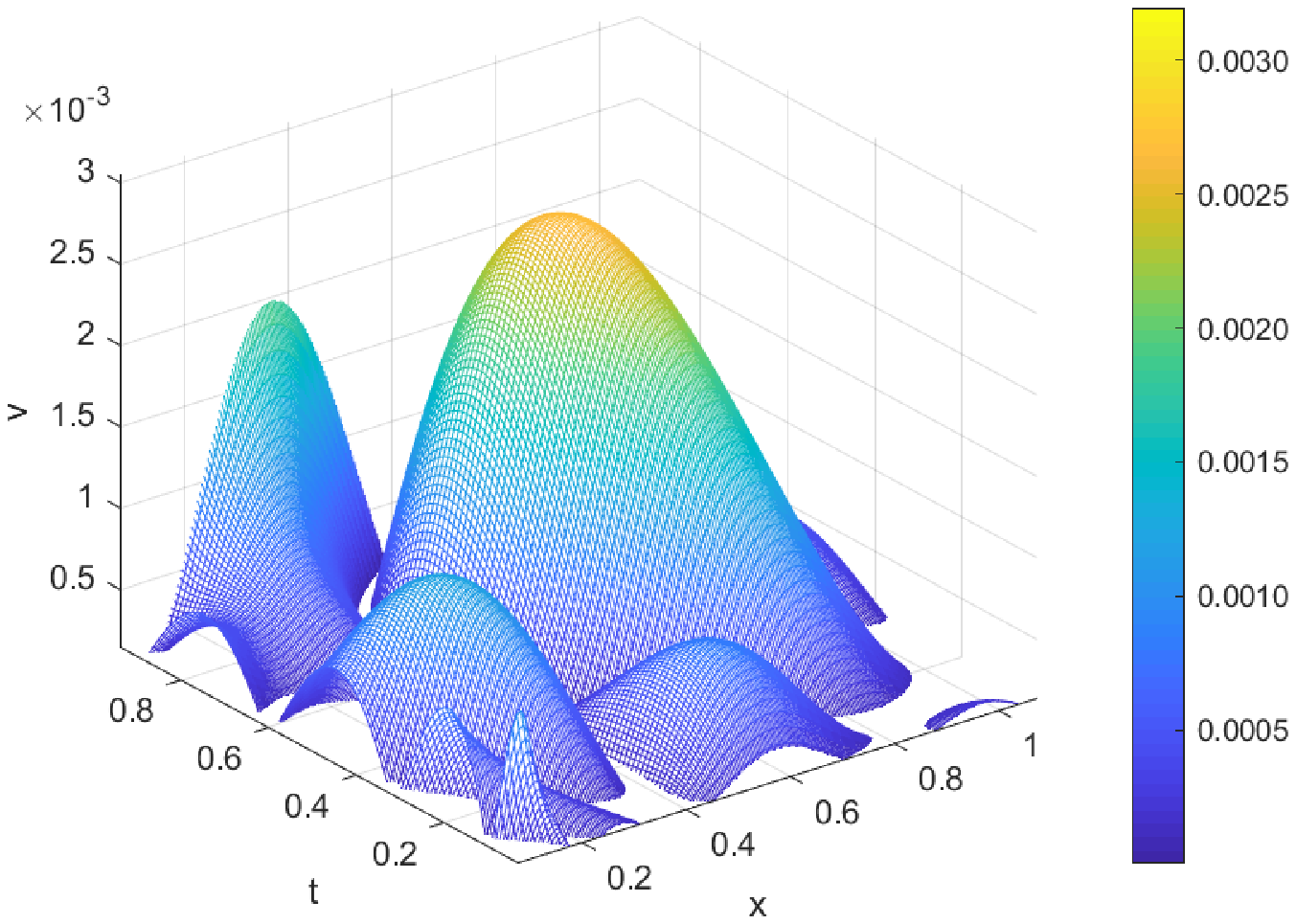}}
		\centerline{D: PINN$_v$}
	\end{minipage}
	\begin{minipage}{0.33\linewidth}
		\vspace{3pt}
		\centerline{\includegraphics[width=\textwidth]{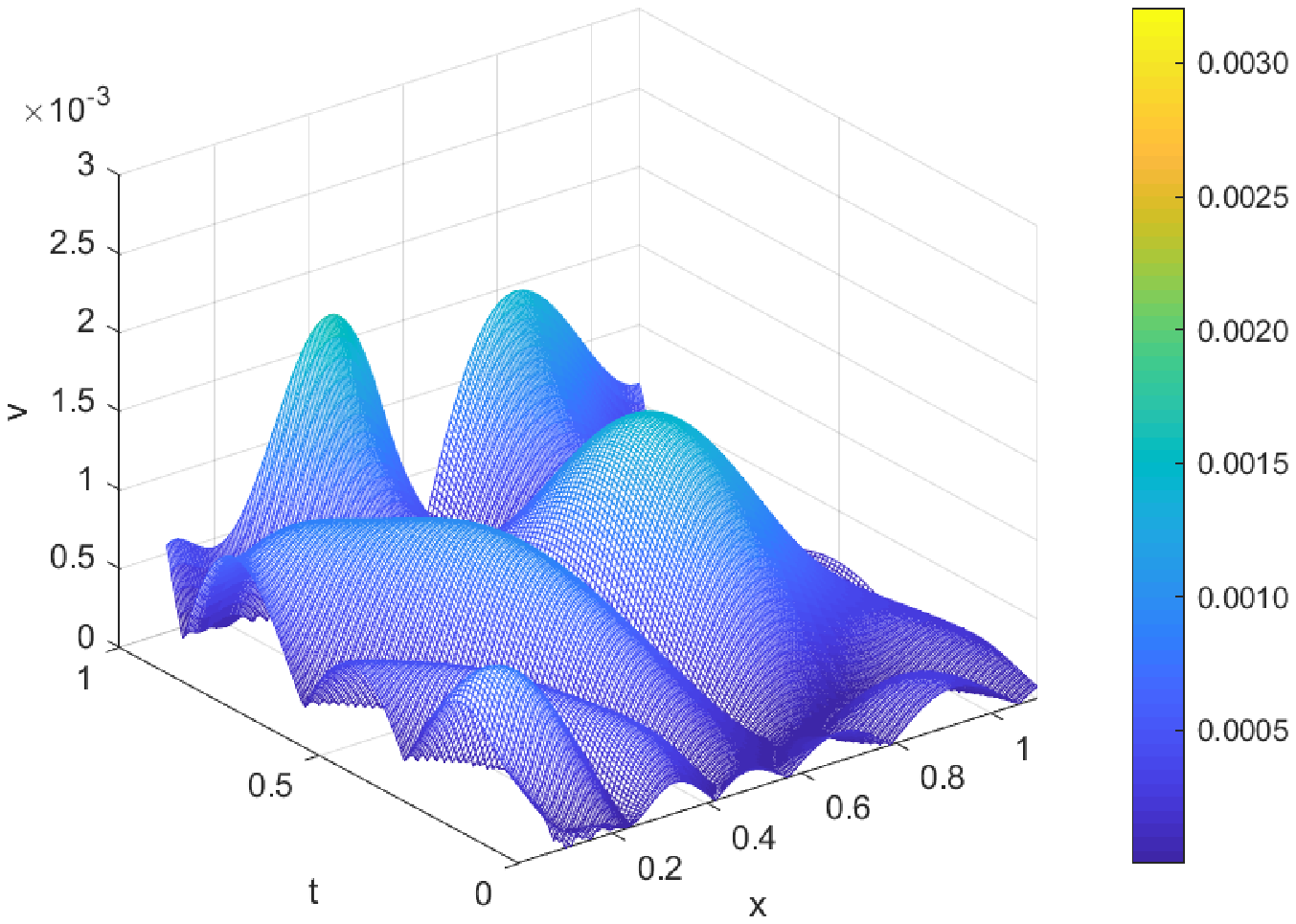}}
		\centerline{E: SPINN$_v$}
	\end{minipage}\\
\begin{minipage}{0.3\linewidth}
		\vspace{3pt}
		\centerline{\includegraphics[width=\textwidth]{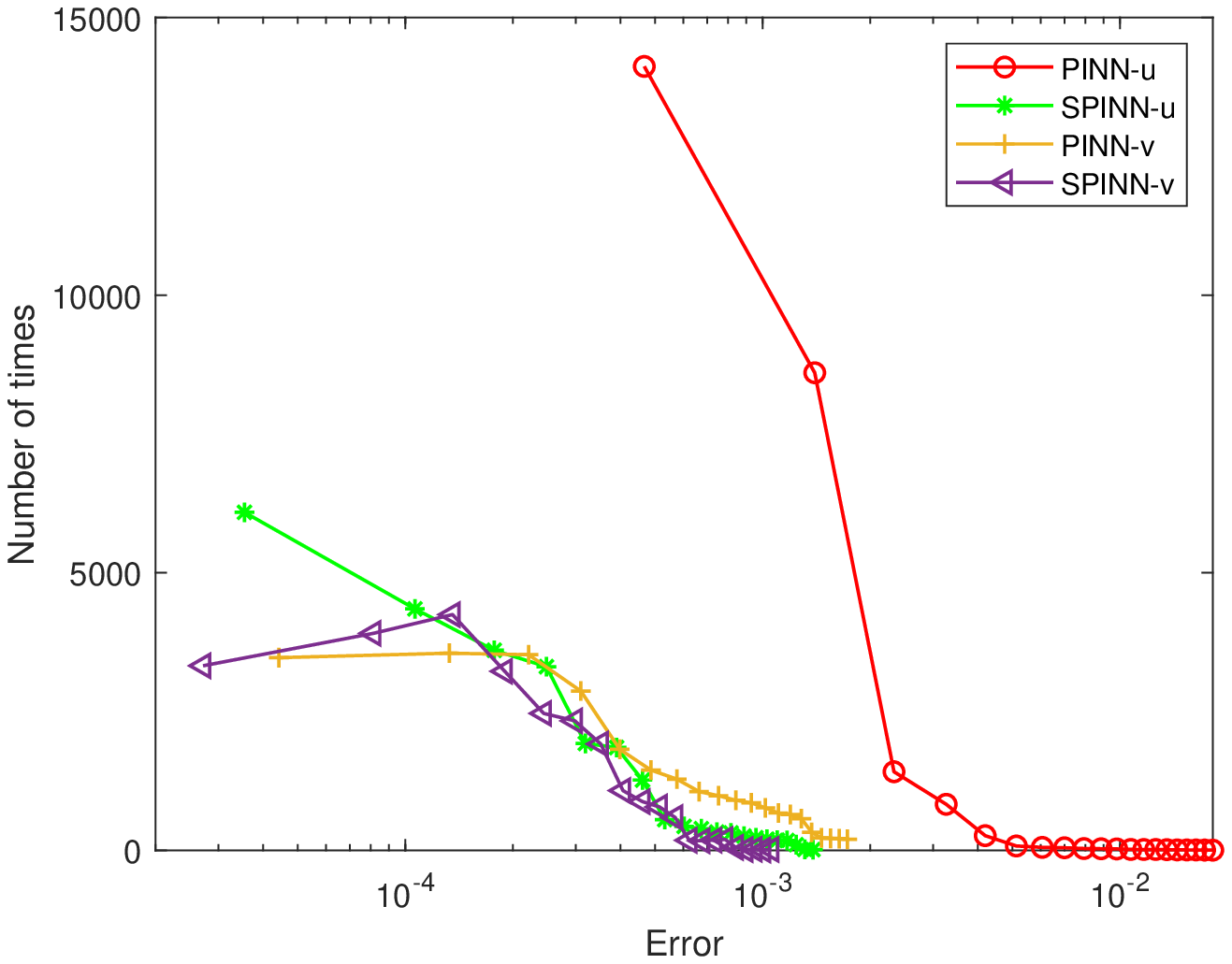}}
		\centerline{F}
	\end{minipage}
	\begin{minipage}{0.33\linewidth}
		\vspace{3pt}
		\centerline{\includegraphics[width=\textwidth]{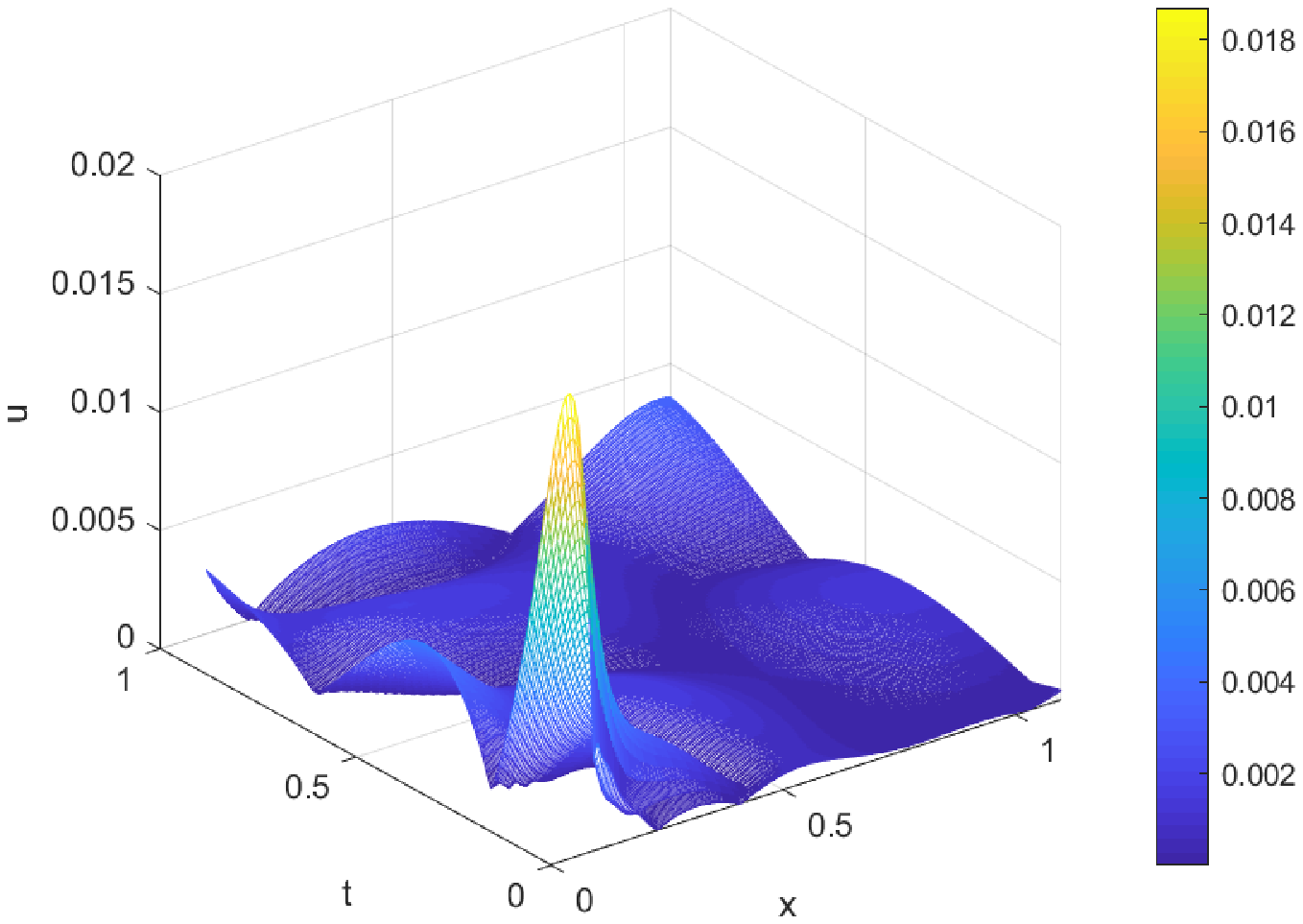}}
		\centerline{G: PINN$_u$}
	\end{minipage}
\begin{minipage}{0.33\linewidth}
		\vspace{3pt}
		\centerline{\includegraphics[width=\textwidth]{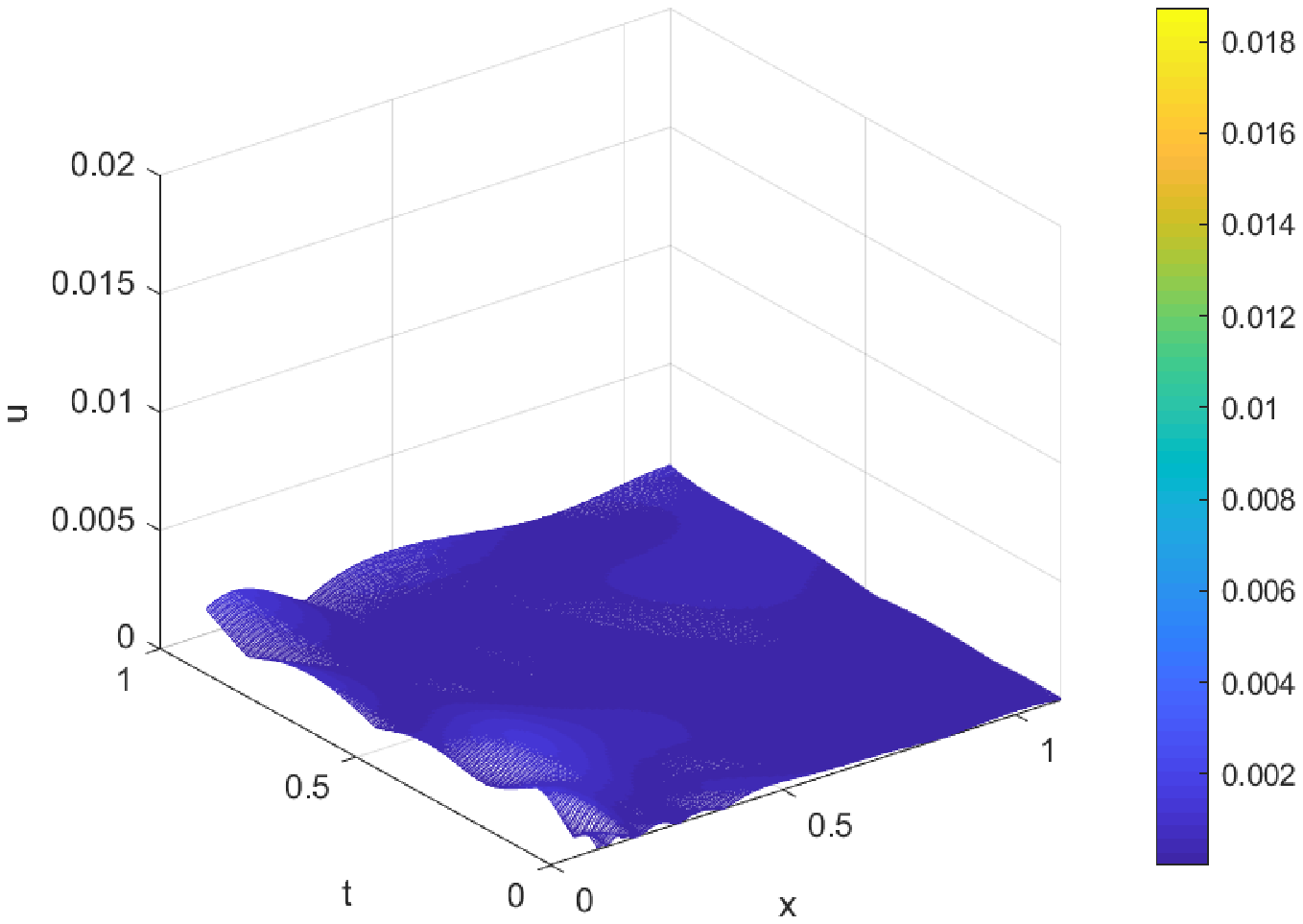}}
		\centerline{I: SPINN$_u$}
	\end{minipage}
\begin{minipage}{0.33\linewidth}
		\vspace{3pt}
		\centerline{\includegraphics[width=\textwidth]{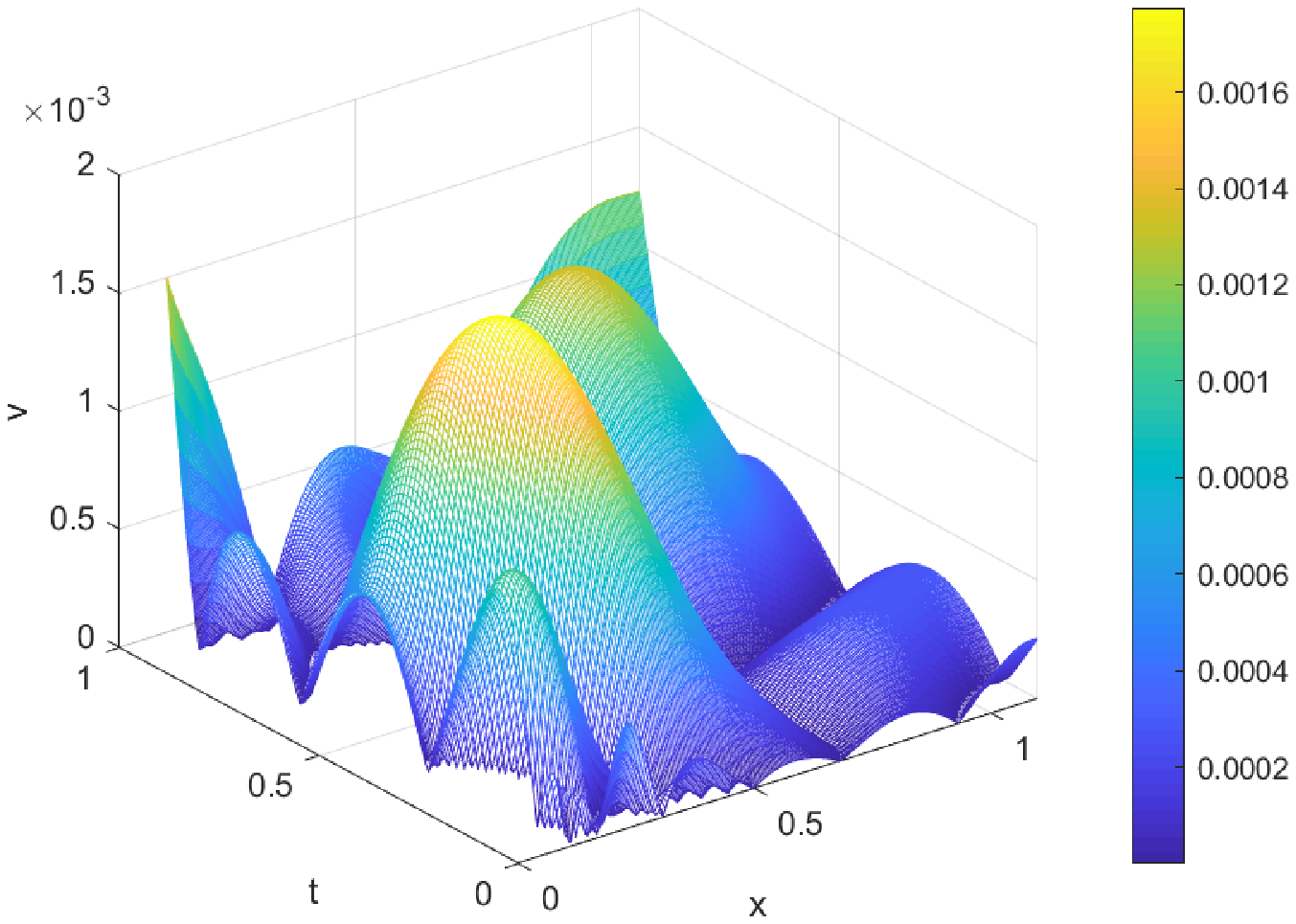}}
		\centerline{H: PINN$_v$}
	\end{minipage}
\begin{minipage}{0.33\linewidth}
		\vspace{3pt}
		\centerline{\includegraphics[width=\textwidth]{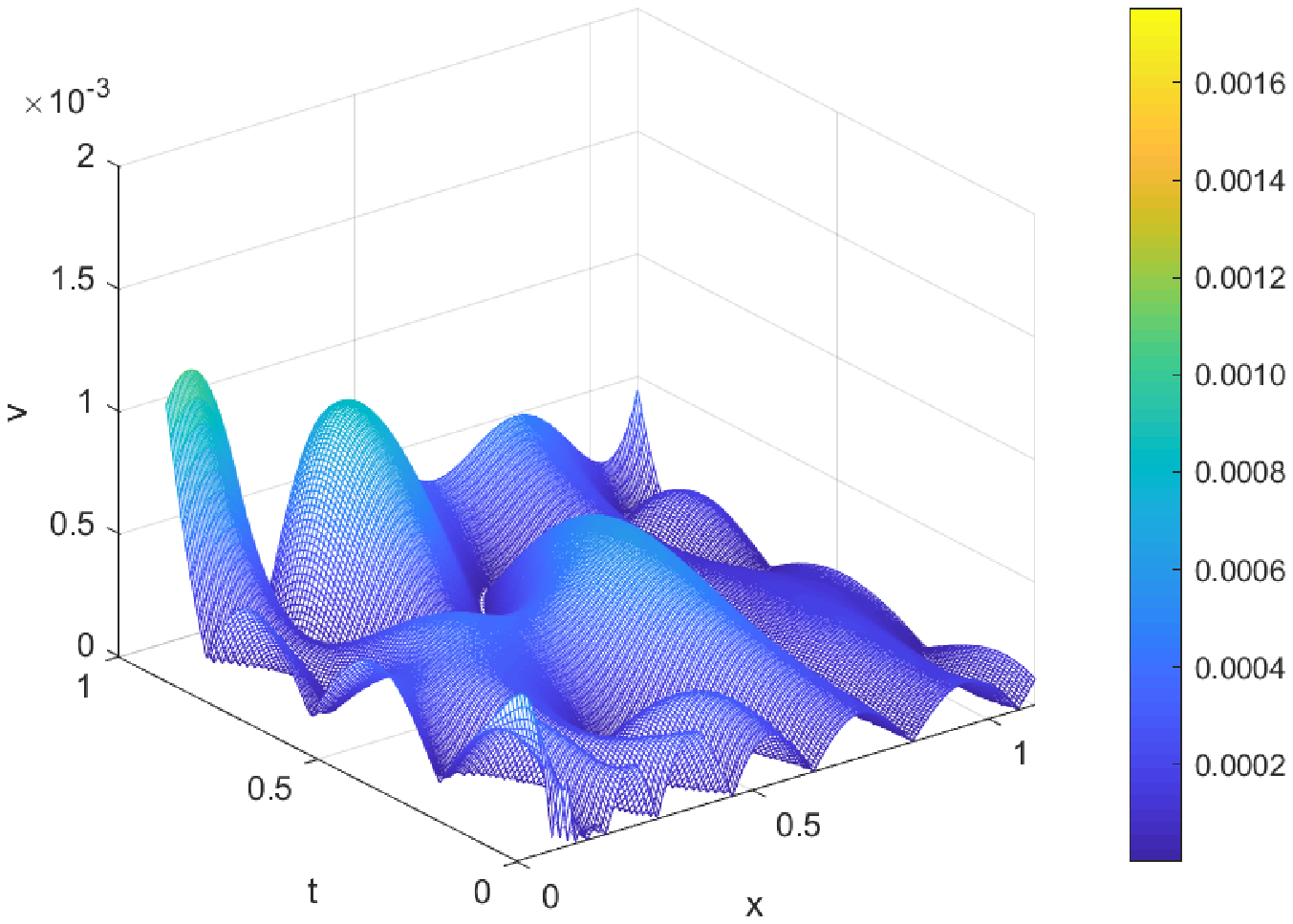}}
		\centerline{J: SPINN$_v$}
	\end{minipage}
	\caption{(Color online) Burgers potential equations: Two graph sets (A,B,C,D,E) and  (F,G,H,I,J) correspond to the two selected bad performances of SPINN in the left and right experiments in Figure \ref{fig1-burgers} respectively. (A and F): Absolute error distributions of $u$ and $v$ for the two bad performances. (B,C,D,E) and (G,H,I,J): Comparisons of absolute errors between PINN and SPINN for the two bad performances of SPINN.}
\label{fig2-burgers}
\end{figure}

Figure \ref{fig2-burgers} shows the absolute errors of the two bad performances for SPINN where the graphes A-E depict for the case of $\widetilde{N}=250$ collocation points while the graphes F-J for the case of 35 neurons. The distributions of absolute errors A and D for the both cases demonstrate that the tendency of peak values of SPINN is more close to zero than PINN, where the improvement of $u$ is more remarkable than $v$.  The pictures of three-dimensional absolute error also expose the effectiveness of SPINN for $u$, approximately flat in the given region, is better than the one of $v$ which has fluctuations in the same region, thus it is possible that the effects of $v$ mainly give rise to the intersections of the shaded areas in Figure \ref{fig1-burgers}.

%
\subsection{Computational cost of SPINN}
\label{sec34}
The above three numerical experiments show that SPINN with the same network structure or collocation points achieves higher accuracy and is more stable than PINN. Since the loss functions in SPINN are added by the ISC, we take for granted that SPINN spends more training time than PINN. However, the final results overturn the idea. We use the relative computational cost of SPINN to PINN, defined by the training time of SPINN divided by the training time of PINN \cite{J}, to quantify the computational overhead of SPINN.  All the computations in this section are performed using the Intel Core i5-11300H CPU.

\begin{figure}[htpb]
	\begin{minipage}{0.5\linewidth}
		\vspace{3pt}
		\centerline{\includegraphics[width=\textwidth]{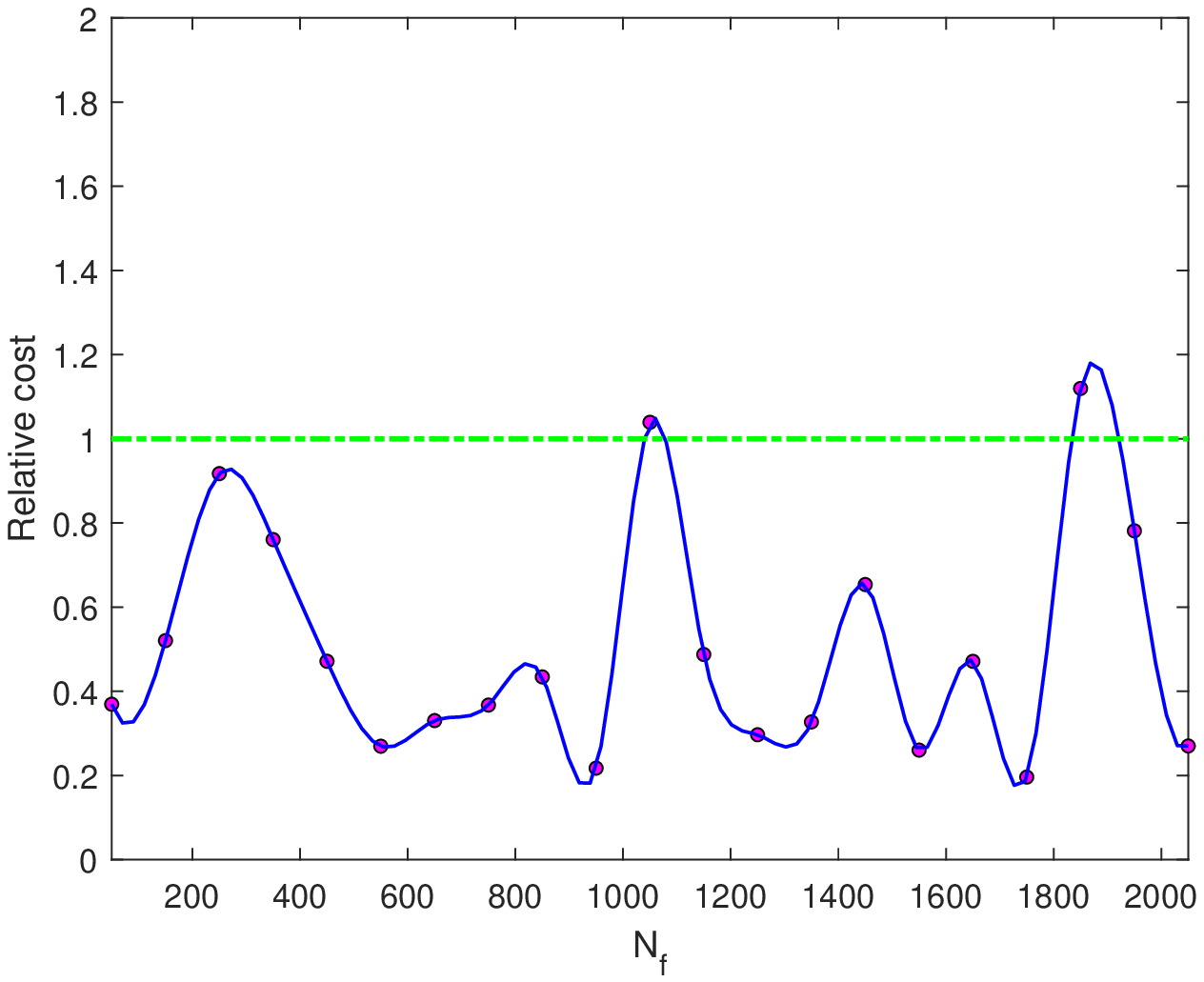}}
        \centerline{A}
	\end{minipage}
	\begin{minipage}{0.5\linewidth}
		\vspace{3pt}
		\centerline{\includegraphics[width=\textwidth]{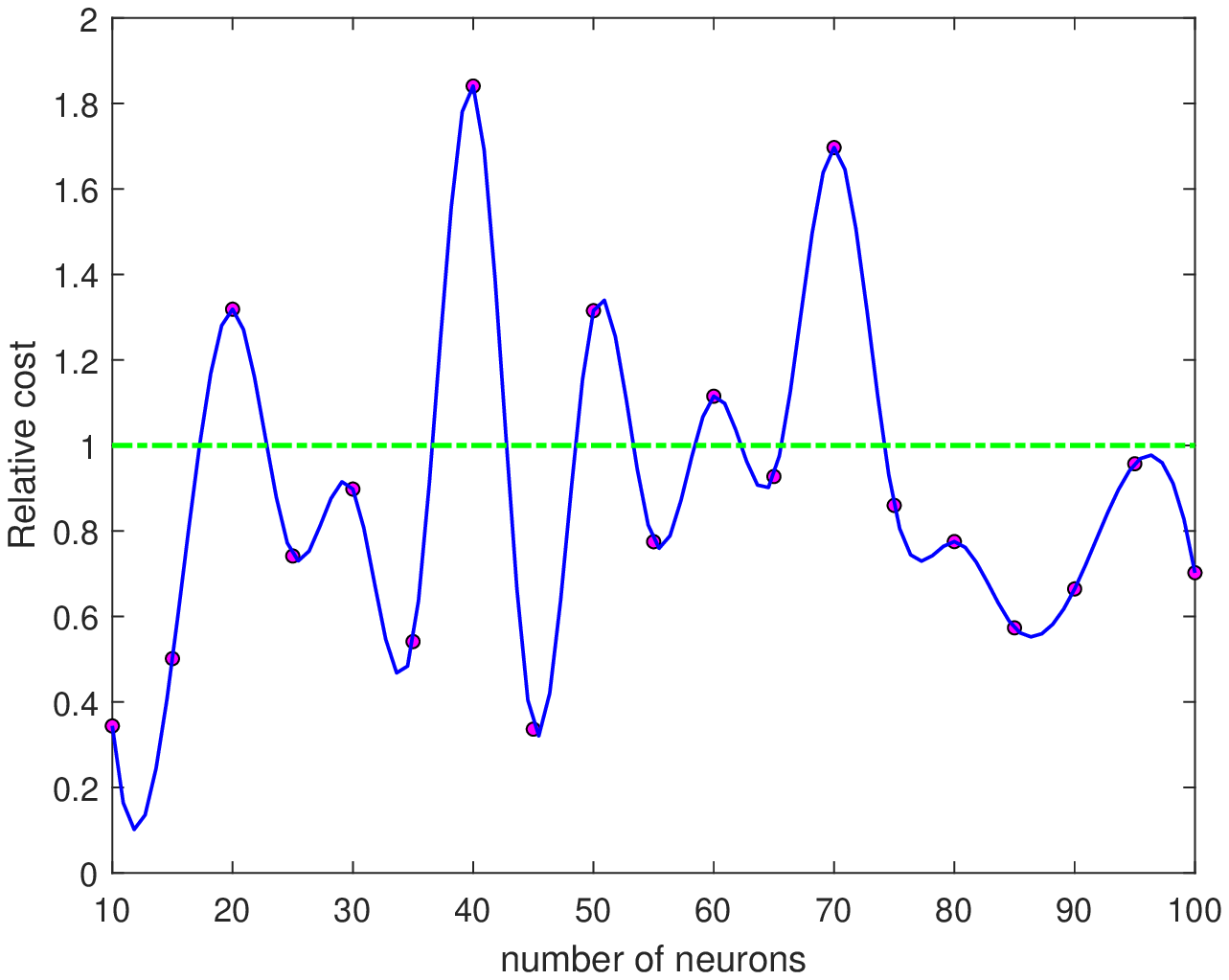}}
        \centerline{B}
	\end{minipage}
	\caption{(Color online) Relative  computational cost distribution of the KDV equation. (A) Keeping the network structure consistent, the number of collocation points varies from 50 to 2050 with step 100. (B) Keeping the number of selected data points unchanged, the numbers of neurons vary from 10 to 100 with step 5.}
\label{fig-cost1}
\end{figure}

For the KdV equation in Section 3.1, the relative computational costs for different numbers of collocation points and neurons in one trial are shown in Figure \ref{fig-cost1}.
In the left graph for the twenty-one collocation points, the relative computational costs below one appear nineteen times, taking 90\%,  the others take 10\% and neither of them exceeded 2. While in the right graph for the nineteen neurons,  the relative computational costs below one emerge fourteen times, taking 74\%,  the others are five times between one and two and take 26\%. It means that SPINN for the KdV equation takes less time than PINN over 70\% cases. Moreover, together with the fact in Figure \ref{fig1-kdv} that SPINN with less collocation points outperforms PINN, we find that SPINN for KdV equation has great advantages than PINN in terms of both accuracy and training time.

\begin{figure}[htp]
	\begin{minipage}{0.5\linewidth}
		\vspace{3pt}
		\centerline{\includegraphics[width=\textwidth]{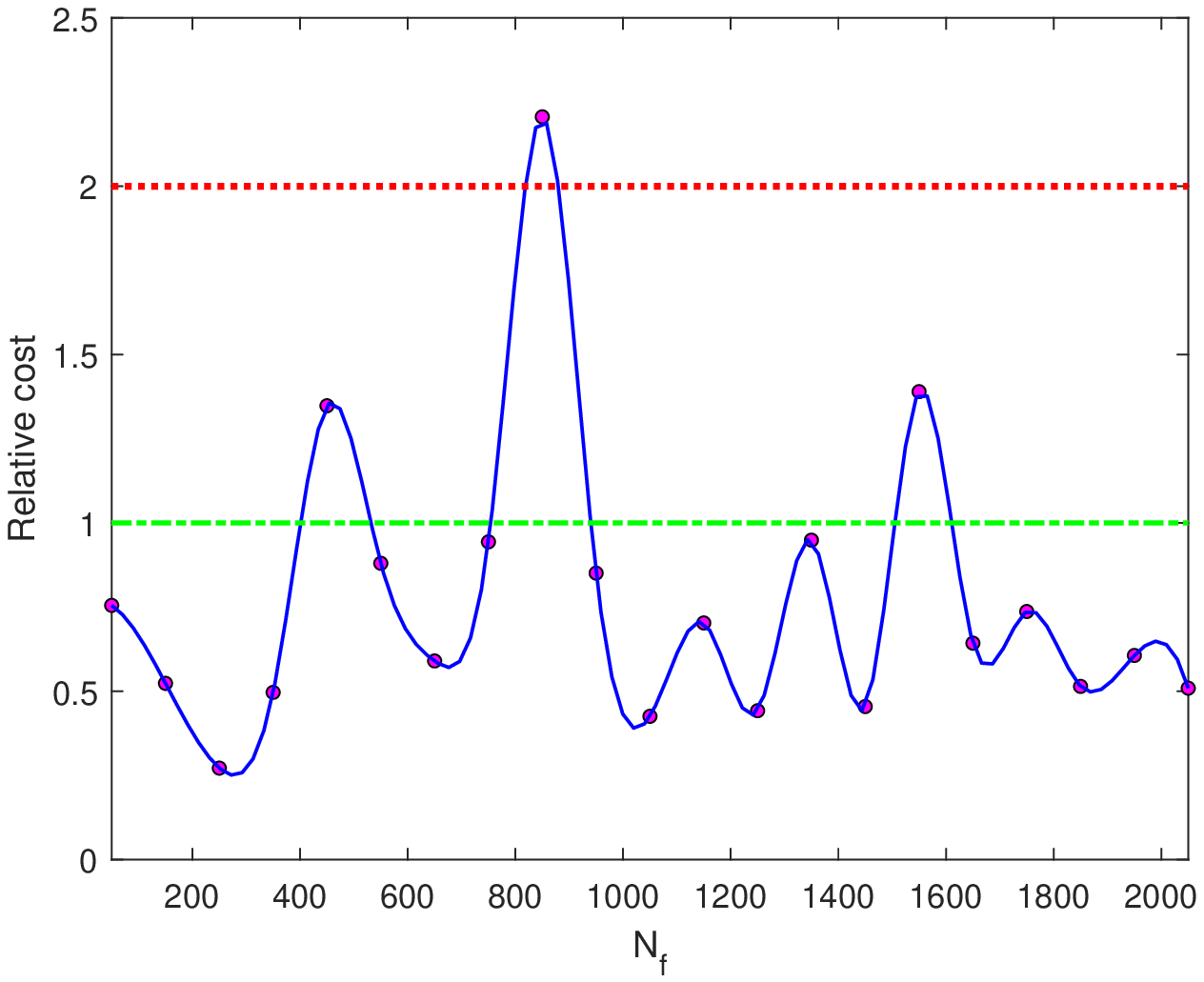}}
        \centerline{A}
	\end{minipage}
	\begin{minipage}{0.5\linewidth}
		\vspace{3pt}
		\centerline{\includegraphics[width=\textwidth]{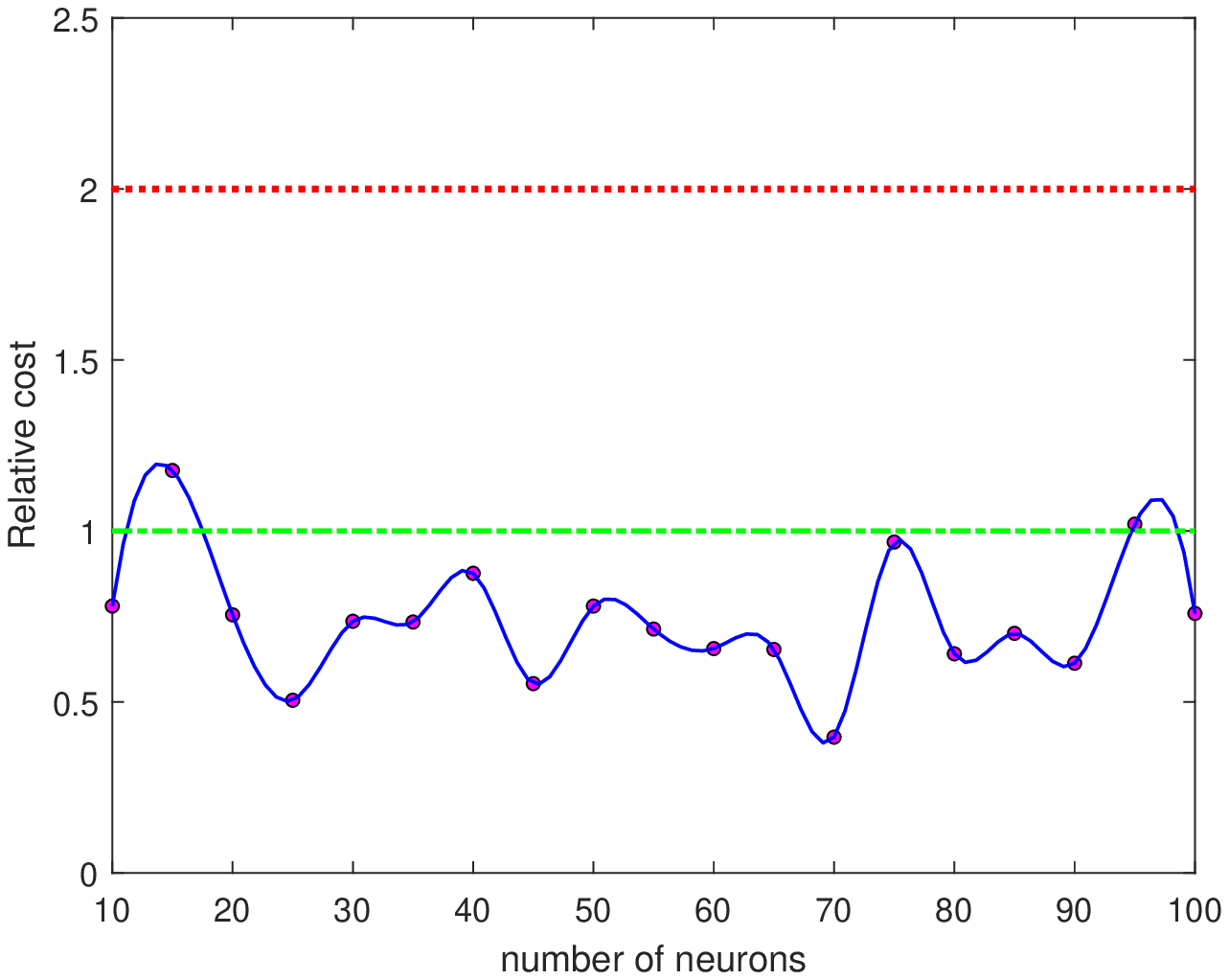}}
        \centerline{B}
	\end{minipage}
	\caption{(Color online) Relative  computational cost distribution of the heat equation. (A) Keeping the network structure consistent, the number of collocation points varies from 500 to 2500 with step 100. (B) Keeping the number of selected data points unchanged, the numbers of neurons vary from 10 to 100 with step 5.}
\label{fig-cost2}
\end{figure}

\begin{figure}[htp]
	\begin{minipage}{0.5\linewidth}
		\vspace{3pt}
		\centerline{\includegraphics[width=\textwidth]{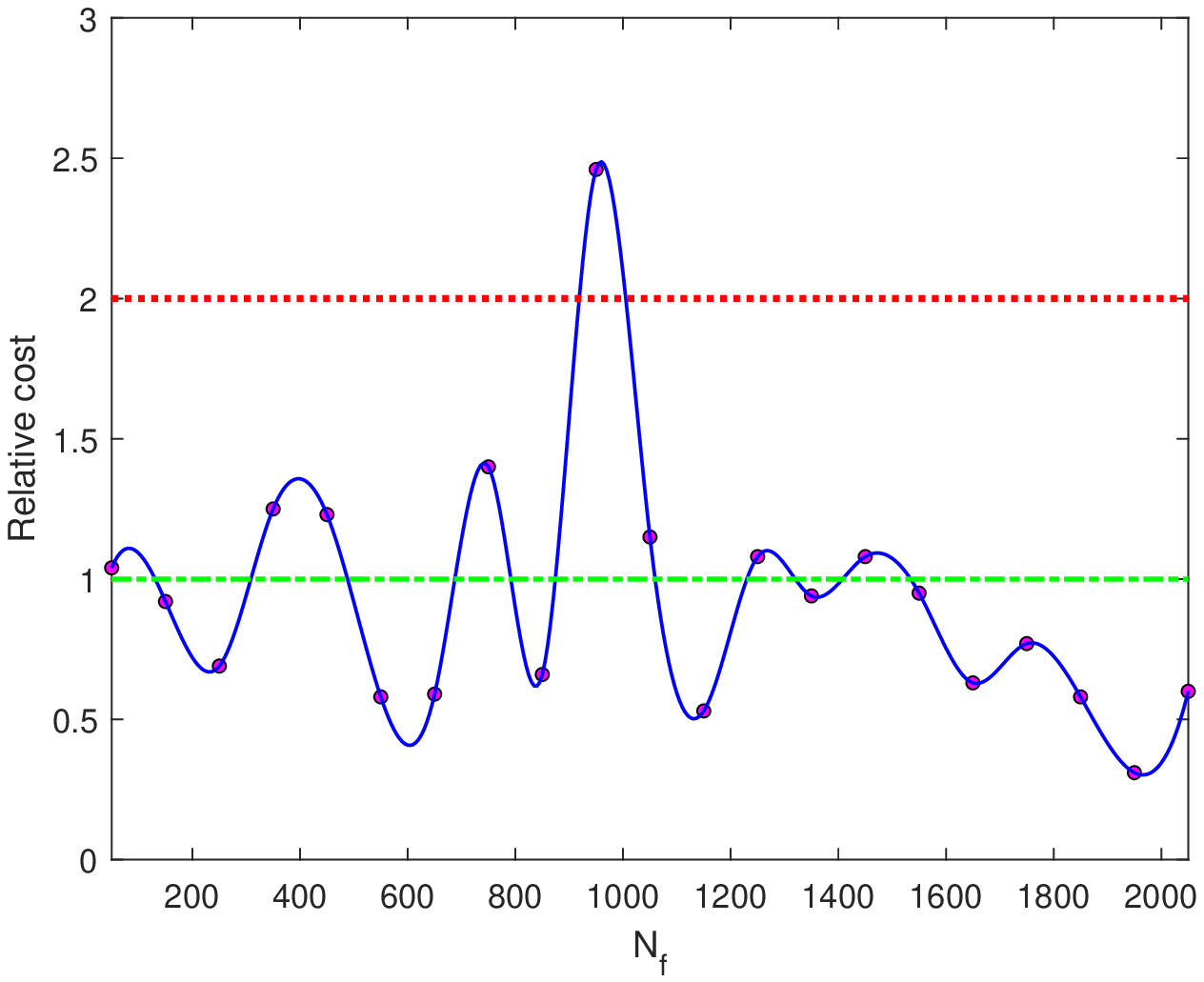}}
        \centerline{A}
	\end{minipage}
	\begin{minipage}{0.5\linewidth}
		\vspace{3pt}
		\centerline{\includegraphics[width=\textwidth]{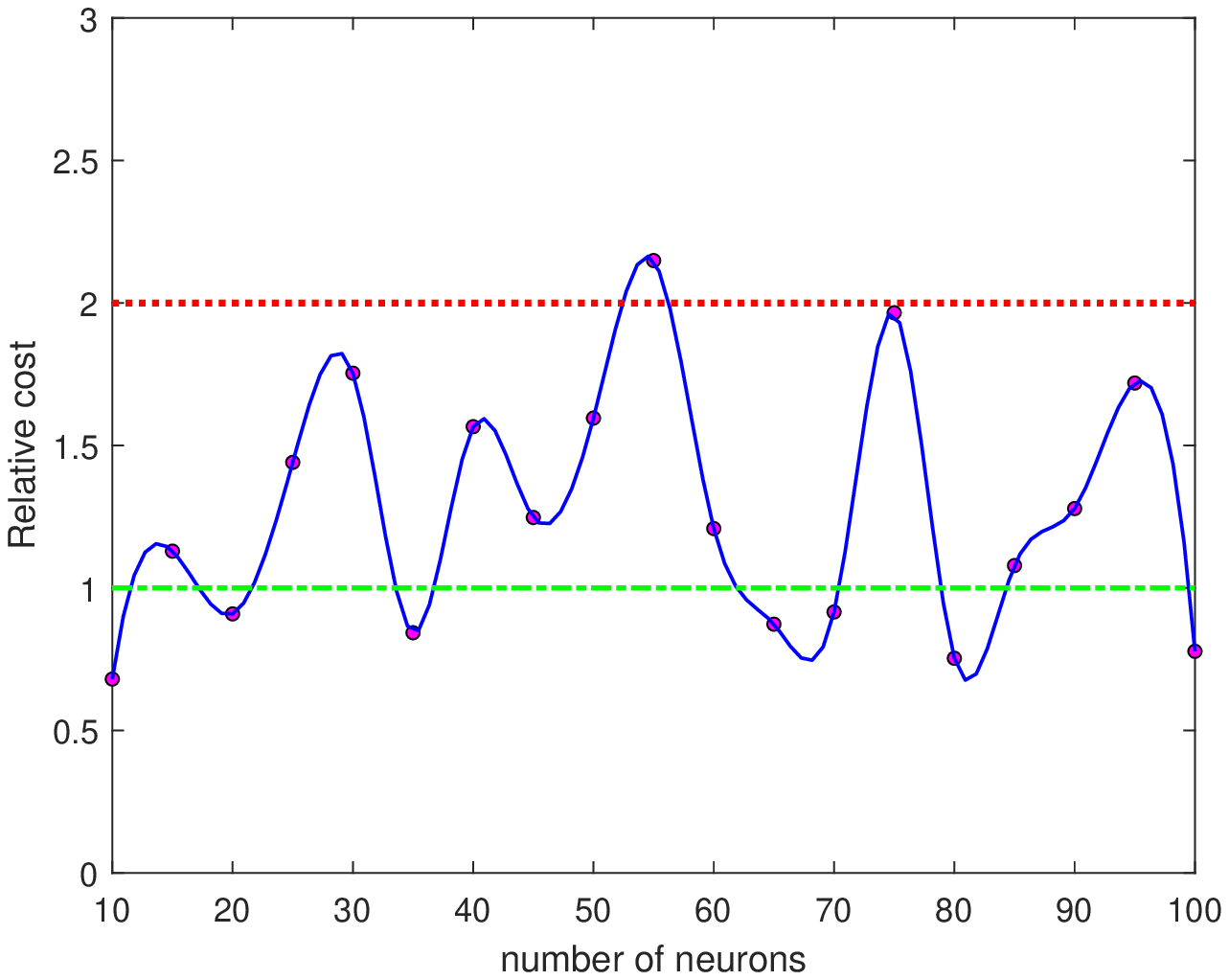}}
        \centerline{B}
	\end{minipage}
	\caption{(Color online) Relative computational cost distribution of the Potential Burgers equations. (A) Keeping the network structure consistent, the number of collocation points varies from 50 to 2050 with step 100. (B) Keeping the number of selected data points unchanged, the numbers of neurons vary from 10 to 100 with step 5.}
\label{fig-cost3}
\end{figure}

The left graph in Figure \ref{fig-cost2} presents the overall same results for the relative computational costs of the Heat equation with twenty one collocations points, 86\% smaller than one and 14\% bigger than one, but only one collocation point surpass two. While in the right graph for the cases of variations of neurons, the relative computational costs for all cases fluctuate around one and do not exceed 1.5.
In Figure \ref{fig-cost3}, the graph A for the collocation points shows the similar results as the Heat equation but the graph B for the variations of neurons presents not good scenario where only seven cases are less than one.

Furthermore, we list the average values of relative computational costs of SPINN to PINN for the above three examples respectively in Table \ref{tab-cost}. The average relative computational costs change between 0.50 to 1.26, and only the case of potential burgers equations with different numbers of neurons surpasses one. Therefore, though an additional loss term ISC is added into the loss function in PINN, SPINN still has huge superiorities than PINN in terms of training time, and thus strongly demonstrate that the inherent symmetry properties of PDEs can accelerate training procedure. 
\begin{table}[htp]
    \centering
    \caption{Relative computational costs of SPINN to PINN for the three examples} \label{si}
    \begin{tabular}{l c}
        \hline\noalign{\smallskip}
         Examples &Relative computational costs  \\
     \noalign{\smallskip}\hline\noalign{\smallskip}
     3.1 KdV equation($\widetilde{N}$) &0.50\\
     3.1 KdV equation(neurons)&0.89\\
     3.2 Heat equation($\widetilde{N}$) &0.77\\
     3.2 Heat equation(neurons)&0.74\\
     3.3 Potential Burgers equations($\widetilde{N}$)&0.93\\
     3.3 Potential Burgers equations(neurons)&1.26\\
     \noalign{\smallskip}\hline
   \end{tabular}
    \label{tab-cost}
\end{table}

\section{Inverse problem of PDEs via SPINN}
We use SPINN method to learn the parameters $\lambda_1$ and $\lambda_2$ in the Burgers equation taking in the potential form
\begin{eqnarray}\label{burpot}
&& f:=u_t-\lambda_1u_{x}^2-\lambda_2u_{xx}=0,~~~~~ x\in[0,2],~~t\in[0.1,1.1],
\end{eqnarray}
which is connected with the celebrated Burgers equation $v_t+vv_x+v_{xx}=0$ by first differentiating Eq.(\ref{burpot}) and then using $v=u_x$ \cite{burger-1948,olv}. 
Eq.\eqref{burpot} is admitted by the symmetry $\mathcal {X}_{burgers}=4\lambda_1t^2\partial_t+4\lambda_1 t x\partial_x-(x^2+2\lambda_2 t)\partial_u$ which gives an exact solution \cite{bai-2022}
\begin{eqnarray}\label{sol-bur}
&& u=\frac{\lambda_2}{\lambda_1} \ln \left(\frac{\lambda_1 x}{\lambda_2 t}+c_1\right)-\frac{\lambda_2}{2 \lambda_1}\ln t-\frac{x^2}{4 \lambda_1 t}+c_2,
\end{eqnarray}
and generates the ISC $g:=4\lambda_1t^2u_t+4\lambda_1 t xu_x+x^2+2\lambda_2 t=0$, where $c_1$ and $c_2$ are two arbitrary constants. In what follows, we choose $\lambda_1=1$ and $\lambda_2=2$ as an example to illustrate the effectiveness of SPINN in learning the parameters $\lambda_i\,(i=1,2)$ as well as the solution.
\subsection{Numerical experiment}
We discretize the solution $u(t,x)$ (\ref{sol-bur}) corresponding to $\lambda_1=1$ and $\lambda_2=2$ into $256\times100$ data points where the spatial region $x\in[0,2]$ and the time region $t\in[0.1,1.1]$ are divided into $N_x=256$ and
$N_t=100$ discrete equidistance points respectively. Then we use the L-BFGS algorithm to optimize the parameters to minimize the loss function
\begin{eqnarray}\label{burgers-mse}
&& MSE=MSE_p+MSE_f+MSE_g,
\end{eqnarray}
where $MSE_p$ corresponds to the training data $\{t^{i},x^{i},u^i\}_{i=0}^{\widehat{N}}$ on $u(t,x)$ while $MSE_f$ enforces the structure imposed by Eq.\eqref{burpot} and $MSE_g$ penalizes the ISC at a finite set of collocation points whose number and location are the same as the training data,
\begin{eqnarray}
&&\no MSE_p=\frac{1}{\widehat{N}}\sum_{i=1}^{\widehat{N}}| u(t^i,x^{i})-u^i|^2, \\
&& \no MSE_f=\frac{1}{\widehat{N}}\sum_{i=1}^{\widehat{N}}|f(t^i,x^{i})|^2,~~~MSE_g=\frac{1}{\widehat{N}}\sum_{i=1}^{\widehat{N}}|g(t^i,x^{i})|^2.
\end{eqnarray}

\begin{figure}[htp]
\centerline{\includegraphics[width=0.9\textwidth]{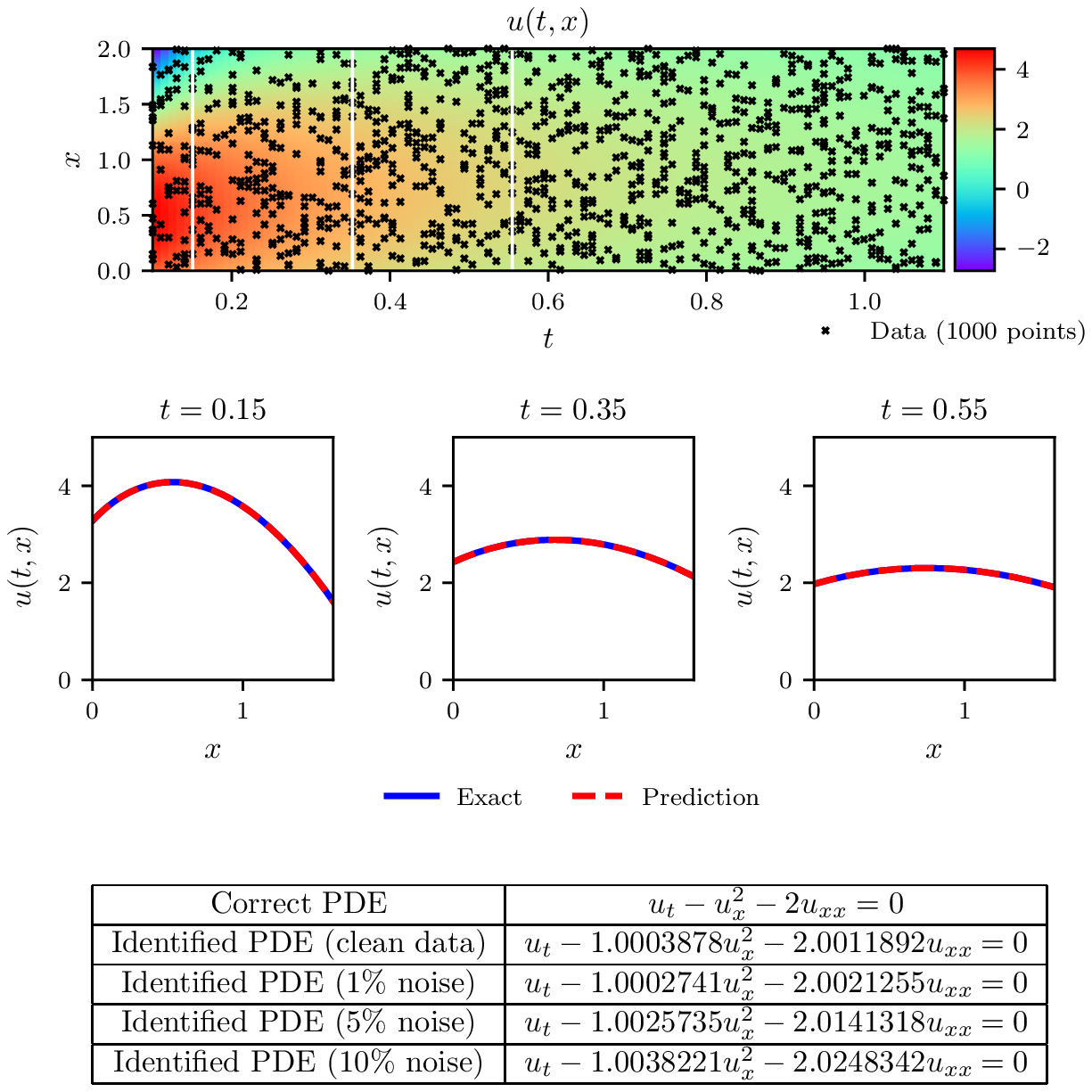}}
\caption{ Burgers equation: Top: Predicted solution $u(t,x)$ along with the training data with $10\%$ noise. Middle: Comparison of the predicted and exact solutions corresponding to the three temporal snapshots depicted by the dashed vertical lines in the top panel. Bottom: Correct PDEs along with the identified
ones obtained by learning $\lambda_1$ and $\lambda_2$.}
\label{bur-inv}
\end{figure}

We randomly sample $N=1000$ training points from the entire spatio-temporal domain $[0,2]\times[0.1,1.1]$ whose locations are listed in the top panel in Figure \ref{bur-inv} to train a 5-layer deep neural network with 40 neurons per hidden layer for predicting the solution $u(t, x)$ as well as the unknown parameters $\lambda_1$ and $\lambda_2$, where the loss function is defined by (\ref{burgers-mse}). The physical propagation diagram of the exact solution and the predicted
solution is shown in the middle panel while the bottom panel shows correct Burgers equation along with the identified ones obtained by the learned $\lambda_1$ and $\lambda_2$.

To further scrutinize the performance of SPINN, we perform a systematic experiment with respect to the number of training data, the noise corruption levels, and the neural network architecture.  Under the $\widehat{N}=1000$ noise-free training points, Table \ref{bur-tab1} shows that the percentage errors in $\lambda_1$ and $\lambda_2$ via SPINN almost outperform the ones by PINN, even three order of magnitude improvement in the case of 8 hidden layer with 10 neurons per layer. In particular, by SPINN the percentage errors in both $\lambda_1$ and $\lambda_2$ with 10 neurons along with 2 hidden layers outperforms the results by PINN with 20 and 40 neurons, which means SPINN with simple architecture can also achieve higher accuracy than PINN with complex one for solving the inverse problem of PDEs.


Meanwhile, by the architecture of 4 hidden layers and 40 neurons per layer, Table \ref{bur-tab2} shows the comparisons of percentage errors in $\lambda_1$ and $\lambda_2$ for the numbers of training data $N_u$ varying from 300 to 1000 and the noise levels from $0\%$ to $10\%$. All the experiment results for SPINN take the obvious superiority than PINN except for the two cases of 700 and 1000 training points and $1\%$ noise where percentage errors in $\lambda_2$ of SPINN is less than PINN but the ones in $\lambda_1$ of SPINN outperform PINN. The reason for the phenomenon is that the best error of predicted solution $u$ of equation (\ref{burpot}) cannot assure both of errors in $\lambda_1$ and $\lambda_2$ better, but at least one of them performs better than another.  The results in Table \ref{bur-tab2} further demonstrate that SPINN method has better robustness than PINN with respect to noise levels in the data, and yields a reasonable identification accuracy even for $10\%$ noise.

\begin{table}[htp]\scriptsize
\captionsetup{width=.9\textwidth,font={scriptsize}}
\caption{Burgers equation: Comparisons of percentage errors in $\lambda_1$ and $\lambda_2$ for different number of hidden layers and neurons per layer. The training data is noise-free and its numbers are $\widehat{N}=1000$.}
\centering
\renewcommand{\arraystretch}{1.2}
\begin{tabular}{c| c ccc c ccc}
\hline
Neurons&\multicolumn{3}{c}{10}&\multicolumn{3}{c}{20}&\multicolumn{2}{c}{40}\\ \hline
\diagbox{Layers}{Errors in $\lambda_1$}{Methods} &SPINN&PINN& &SPINN&PINN& &SPINN&PINN \\ \hline
2&{0.1231}&{0.2752}& &{0.0744}&{1.3101}& &{0.0152}&{0.4464}\\
4&{0.1025}&{0.1485}&&{0.0326}&{0.8435}&&{0.0388}&{0.8598}\\
6&{0.6462}&{0.8165}&&{0.1223}&{1.4636}&&{0.0055}&{0.2801}\\
8&{0.0422}&{10.2716}&&{0.2191}&{2.2273}&&{0.3140}&{0.8458}\\ \hline
\end{tabular}
\begin{tabular}{c| c ccc c ccc}
\hline
Neurons&\multicolumn{3}{c}{10}&\multicolumn{3}{c}{20}&\multicolumn{2}{c}{40}\\ \hline
\diagbox{Layers}{Errors in $\lambda_2$}{Methods} &SPINN&PINN& &SPINN&PINN& &SPINN&PINN \\ \hline
2&0.1356&0.8227& &0.1001&0.2036& &0.0222&0.2106\\
4&0.1358&0.4748&&0.0528&0.1151&&0.0595&0.2627\\
6&0.9693&0.4942&&0.1490&0.2049&&0.0353&0.1413\\
8&0.1524&5.6987&&0.2300&0.4987&&0.4407&0.1148\\ \hline
\end{tabular}
\label{bur-tab1}
\end{table}

\begin{table}[htp]\scriptsize
\captionsetup{width=.9\textwidth,font={scriptsize}}
\caption{Burgers equation: Comparisons of percentage errors in $\lambda_1$ and $\lambda_2$ for different number of training data $N_u$ corrupted by different noise levels. The neural network architecture is fixed to 4 hidden layers and 40 neurons per layer.}
\centering
\renewcommand{\arraystretch}{1.2}
\begin{tabular}{c| c ccc c c ccc}
\hline
Noise&\multicolumn{2}{c}{0\%}&\multicolumn{2}{c}{1\%}&\multicolumn{2}{c}{5\%}&\multicolumn{2}{c}{10\%}\\ \hline
\diagbox{$N_u$}{Errors in $\lambda_1$}{Methods} &SPINN&PINN& SPINN&PINN& SPINN&PINN& SPINN&PINN \\ \hline
300&{0.0527}&{0.5824}& {0.0449}&{0.0737}& {0.4424}&{2.3174}&{0.9203}&{5.2090}\\
500&{0.0424}&{1.4337}&{0.0246}&{1.1675}&{0.4981}&{1.4893}&{0.9246}&{5.3572}\\
700&{0.0002}&{1.2396}&{0.0868}&{0.8315}&{0.2734}&{1.5816} &{0.5821}&{4.1576}\\
1000&{0.0388}&{0.8598}&{0.0274}&{0.6913}&{0.2573}&{0.9094} &{0.3822}&{2.5409}\\ \hline
\end{tabular}
\begin{tabular}{c| c ccc c ccc c}
\hline
Noise&\multicolumn{2}{c}{0\%}&\multicolumn{2}{c}{1\%}&\multicolumn{2}{c}{5\%}&\multicolumn{2}{c}{10\%}\\ \hline
\diagbox{$N_u$}{Errors in $\lambda_2$}{Methods} &SPINN&PINN& SPINN&PINN& SPINN&PINN& SPINN&PINN \\ \hline
300&{0.0948}&{0.3137}& {0.2443}&{0.2628}& {1.5120}&{2.7760}& {3.0701}&{5.9800}\\
500&{0.0457}&{0.5583}&{0.1751}&{0.3413}&{1.2612}&{1.6167}& {2.5189}&{3.8045}\\
700&{0.0229}&{0.2460}&{0.1892}&{0.0105}&{0.7619}&{0.9137}&{1.6237}&{2.4658}\\
1000&{0.0595}&{0.2627}&{0.1063}&{0.0665}&{0.7066}&{0.7922}& {1.2417}&{2.1771}\\ \hline
\end{tabular}
\label{bur-tab2}
\end{table}

\subsection{Relative computational cost}
\begin{figure}[htp]
	\begin{minipage}{0.5\linewidth}
		\vspace{3pt}
		\centerline{\includegraphics[width=\textwidth]{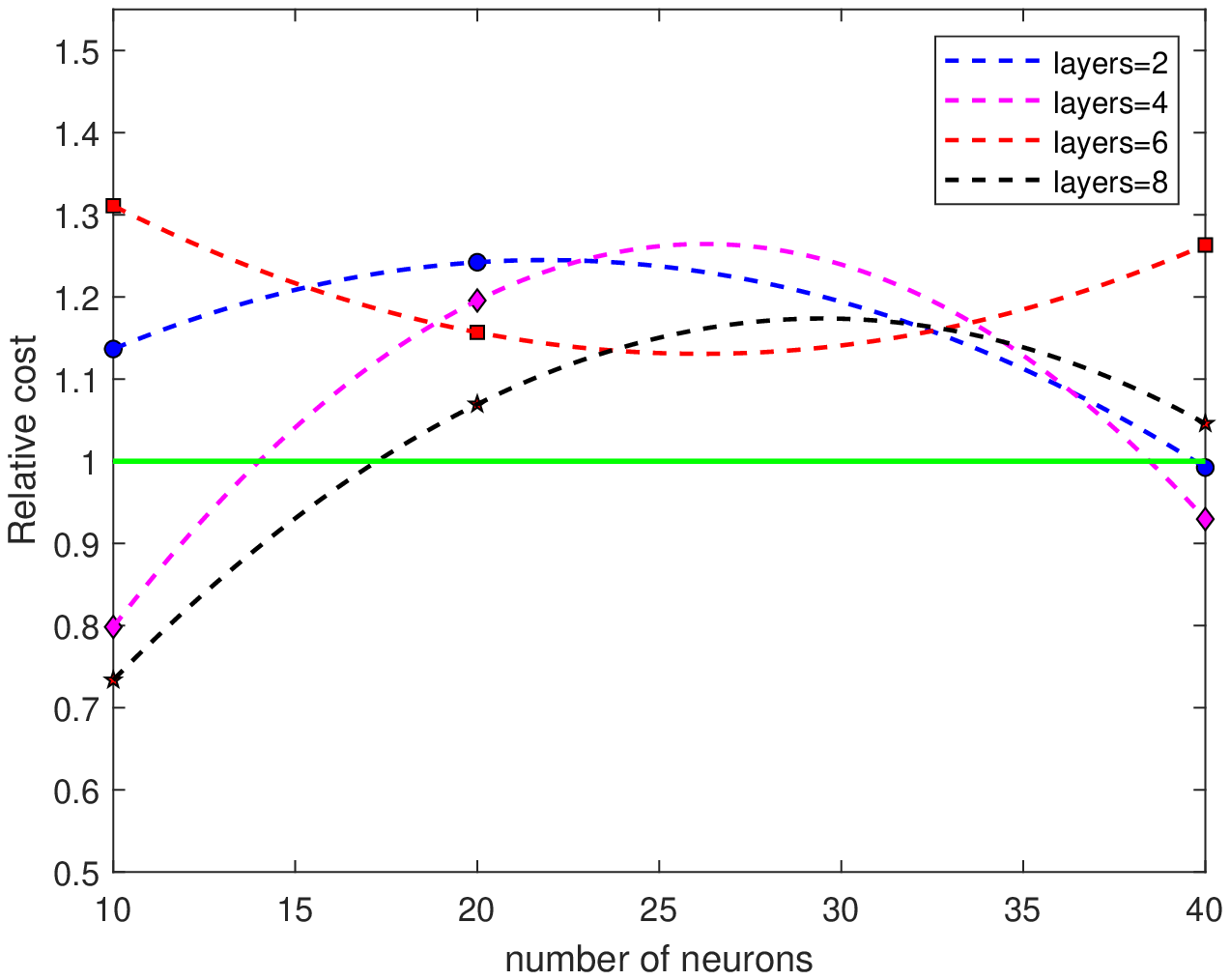}}
        \centerline{A}
	\end{minipage}
	\begin{minipage}{0.5\linewidth}
		\vspace{3pt}
		\centerline{\includegraphics[width=\textwidth]{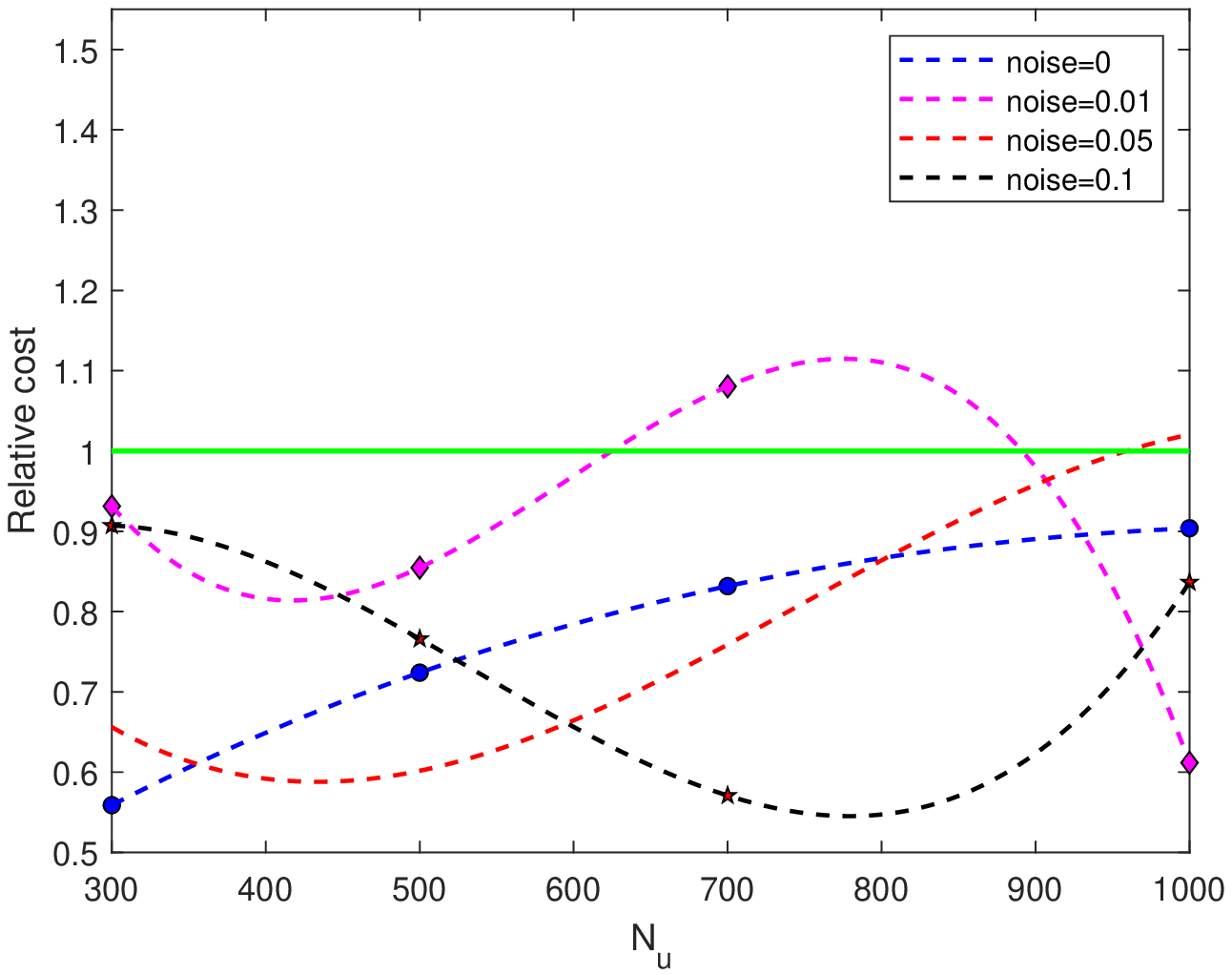}}
        \centerline{B}
	\end{minipage}
	\caption{(Color online) Relative computational cost distribution of the Burgers equations. (A) Keeping the network structure consistent, the number of collocation points varies from 300 to 1000 and the noise levels increase from $0\%$ to $10\%$. (B) Keeping the number of selected data points unchanged, the numbers of neurons vary from 10, 20 to 40 and the numbers of layers increase from 2 to 8.}
\label{fig-costdyn}
\end{figure}

We now turn to the training time of SPINN in dealing with the inverse problem of PDEs and still use the relative computational cost defined in Subsection \ref{sec34}, i.e. the training time of SPINN divided by the one of PINN, to check the influence of the new added ISC in the loss function of SPINN. All the performances are done with the same seed and on the computer with Intel Core i5-11300H CPU. In Figure \ref{fig-costdyn}, the graph A shows the relative computational costs in Table \ref{bur-tab2} where most of the relative computational costs are bigger than one but less than two, which means that penalizing the ISC in the loss function of SPINN takes more time than its roles in accelerating the procedure of prescribed solution approximating exact solution. While the graph B describes the relative computational costs in Table \ref{bur-tab1} and shows that the four relative computational costs are almost all less than 1, which means that the training time of SPINN is less than PINN, except that the case of $1\%$ noise gets the peak at about 750 collocation points and then drops off.

\section{Conclusion}
In this paper, we propose a new SPINN, which incorporates the ISC induced by Lie symmetry or non-classical symmetry into the loss function of PINN, to improve the accuracy and reliability of solutions of PDEs. Numerical experiments for the forward and inverse problems of PDEs show that the proposed SPINN  clearly outperforms PINN where the $L_2$ relative error, error reduction rate as well as the absolute error are used to test the effectiveness. The merits of SPINN for both the forward and inverse problems of PDEs are formulated as follows:

I). SPINN with less training points and simple network architecture can reach a high accuracy value which can not be obtained by PINN within the tested intervals.

Specifically, for the forward problem, the neural network architecture used in the experiments are comparably simple, 3-layer in KdV and potential Burgers equation and 4-layer in Heat equation, but the $L_2$ relative error of SPINN can reach $10^{-5}$ which is seldom in the literatures and also shows that proper deep neural network can generate high accuracy numerical solutions, and generally has one order of magnitude improvement than PINN, even two or three orders of magnitude. While for the inverse problem, the considered architecture for the Burgers equation is relatively simple than the ones in the literatures, usually at least 8 hidden layers and 20 neurons or twenty thousands of iterations,  but gives more higher accuracies on the parameters as well as the predicted solutions.

II). The relative computational cost of the proposed SPINN to PINN does not increase dramatically, where only few particular cases exceed two times of PINN, while the average relative computational costs keep small increases or decreases.

In addition, since the symmetry is an inherent property of PDEs and its related theories have been developed maturely, thus SPINN has broad application prospects and more strong robustness. Our experimental results further demonstrate that the deep neural network enforced by the inherent physical properties of PDEs such as the symmetry information  is very effective for finding high accuracy numerical solutions of PDEs and is worthy of further exploring, for example,  the collection of SPINN with the gradient-enhanced PINN, inserting the generalized symmetry into the loss function of PINN, etc. Such works are under consideration and will be reported in the future.

\section*{Acknowledgements}
The paper is supported by the Beijing Natural Science Foundation (No. 1222014), the National Natural Science Foundation of China (No. 11671014) and the Cross Research Project for Minzu University of China; R\&D Program of Beijing Municipal Education Commission (Nos. KM202110009006 and KM201910009001).
\\\\
\textbf{Declarations of interest: none}

\end{document}